\documentclass[12pt,a4paper]{article}

\usepackage[utf8]{inputenc}
\usepackage{amsmath, amssymb, amsthm}
\usepackage{graphicx}
\usepackage{hyperref}
\usepackage{geometry}
\usepackage[authoryear]{natbib}   
\usepackage{listings}
\usepackage{xcolor}

\usepackage{float}

\usepackage{placeins}  
\lstdefinestyle{python}{
  language=Python,
  basicstyle=\ttfamily\small,
  keywordstyle=\color{blue},
  commentstyle=\color{gray},
  stringstyle=\color{orange},
  showstringspaces=false,
  breaklines=true,
  frame=single,
  numbers=left,
  numberstyle=\tiny\color{gray},
  tabsize=4
}
\usepackage{url}
\usepackage{hyperref}
\usepackage{booktabs}
\DeclareUnicodeCharacter{2061}{}

\title{ AuON: A Linear-time Alternative to Orthogonal Momentum Updates}
\author{Dipan Maity \\
\texttt{dipanai.xyz@gmail.com}}

\date{}

\begin{document}

\maketitle

\begin{abstract}

\noindent Orthogonal momentum gradient updates have emerged to overcome the limitations of vector-based optimizers like Adam. The vector-based optimizer Adam suffers from high memory costs and ill-conditioned momentum gradient updates. However, traditional Orthogonal momentum approaches, such as SVD/QR decomposition, suffer from high computational and memory costs and underperform compared to well-tuned SGD with momentum. Recent advances, such as Muon, improve efficiency by applying momentum before orthogonalization and approximate orthogonal matrices via Newton-Schulz iterations, which gives better GPU utilization, active high TFLOPS, and reduces memory usage by up to 3x. Nevertheless,  Muon(Vanilla) suffers from exploding attention logits and has cubic computation complexity. In this paper we deep dive into orthogonal momentum gradient updates to find the main properties that help Muon achieve remarkable performance. We propose \textbf{AuON} (Alternative Unit-norm momentum updates by Normalized nonlinear scaling), a linear-time optimizer that achieves strong performance without approximate orthogonal matrices, while preserving structural alignment and reconditioning ill-posed updates. AuON has an automatic (\textbf{"emergency brake"}) to handle exploding attention logits. We further introduce a hybrid variant (\textbf{ Hybrid-AuON})that applies the linear transformations with   Newton-Schulz iterations, which outperforms Muon in the language modeling tasks.
Code is available at: \url{https://github.com/ryyzn9/AuON}

\end{abstract}

\section{Introduction}
\begin{quote}
\emph{"If you want to achieve extraordinary progress in AI, you should enhance the optimizer, as it fundamentally determines how models learn."}
\end{quote}

\noindent The pursuit of computational efficiency in training Large Language Models (LLMs) has catalyzed a shift away from purely scalar, coordinate-wise adaptive optimizers toward geometry-aware methods. For over a decade, the optimization landscape has been hegemonized by first-order adaptive moment estimation algorithms, principally Adam and its decoupled weight-decay variant, AdamW. These methods, while robust, operate under the simplifying assumption that the parameter space is Euclidean and that each parameter can be adapted independently based on its gradient history. However, as neural architectures have scaled into the regime of trillions of parameters—exemplified by models like the Mixture-of-Experts (MoE)  limitations of this coordinate-wise paradigm have become increasingly stark. The inefficiencies manifest not merely in convergence speed but in the disconnect between the optimizer’s update steps and the underlying matrix structure of modern neural networks. Principally, Adam and its decoupled weight-decay variant, AdamW, suffer from ill-conditioning of gradient and momentum updates due to the momentum for a linear layer; naturally, a 2D matrix tends to become almost low-rank in practice. This means that only a small number of dominant directions really drive the updates. Although rare directions seem minor, they are often essential for effective learning and can help capture more nuanced patterns in the data.  Empirically, these momentum updates often exhibit a high condition number, with most of the energy concentrated in a few dominant directions. In practical terms, the update vectors are nearly low-rank: a handful of directions dictate the optimization trajectory while many potentially informative directions may be suppressed. This imbalance reminds us of a squashed ball that can only roll efficiently along a single axis, ignoring other pathways that may be equally important for generalization and representation learning. One solution is to make all update directions unit length; recent work has proposed orthogonalization of gradients and momentum updates to achieve this property. Adam and its variant AdamW have another critical limitation: they require keeping two extra variables for every model parameter, meaning the optimizer state takes up about twice as much memory as the model itself. Furthermore, Adam treats all parameters as a single long vector, updating each value independently without considering any internal structure of the parameters.

To overcome these issues, researchers proposed orthogonalizing the momentum gradient update matrix, which can effectively discard the scaling information encoded in the singular values and modify each direction to enforce perpendicularity, redistributing the update length into unit vectors along different directions. In this sense, the resulting update behaves as a \emph{unit-norm} update in the spectral domain, emphasizing the geometric structure of the optimization landscape rather than the raw gradient magnitudes. In simple terms, orthogonalization amplifies 'rare directions' with small magnitude in the update, but which are nevertheless important for learning. This perspective highlights how orthogonalization can prioritize exploration across all relevant directions, mitigating the dominance of a few high-energy components and facilitating more balanced learning dynamics \cite{zhang2025adagradmeetsmuonadaptive}. Orthogonalized updates can therefore be interpreted as spectral descent directions, ensuring that updates explore the space more evenly-crucial for generalization and representation learning.

\begin{figure}[!htb]
    \centering
    \includegraphics[width=1.0\linewidth]{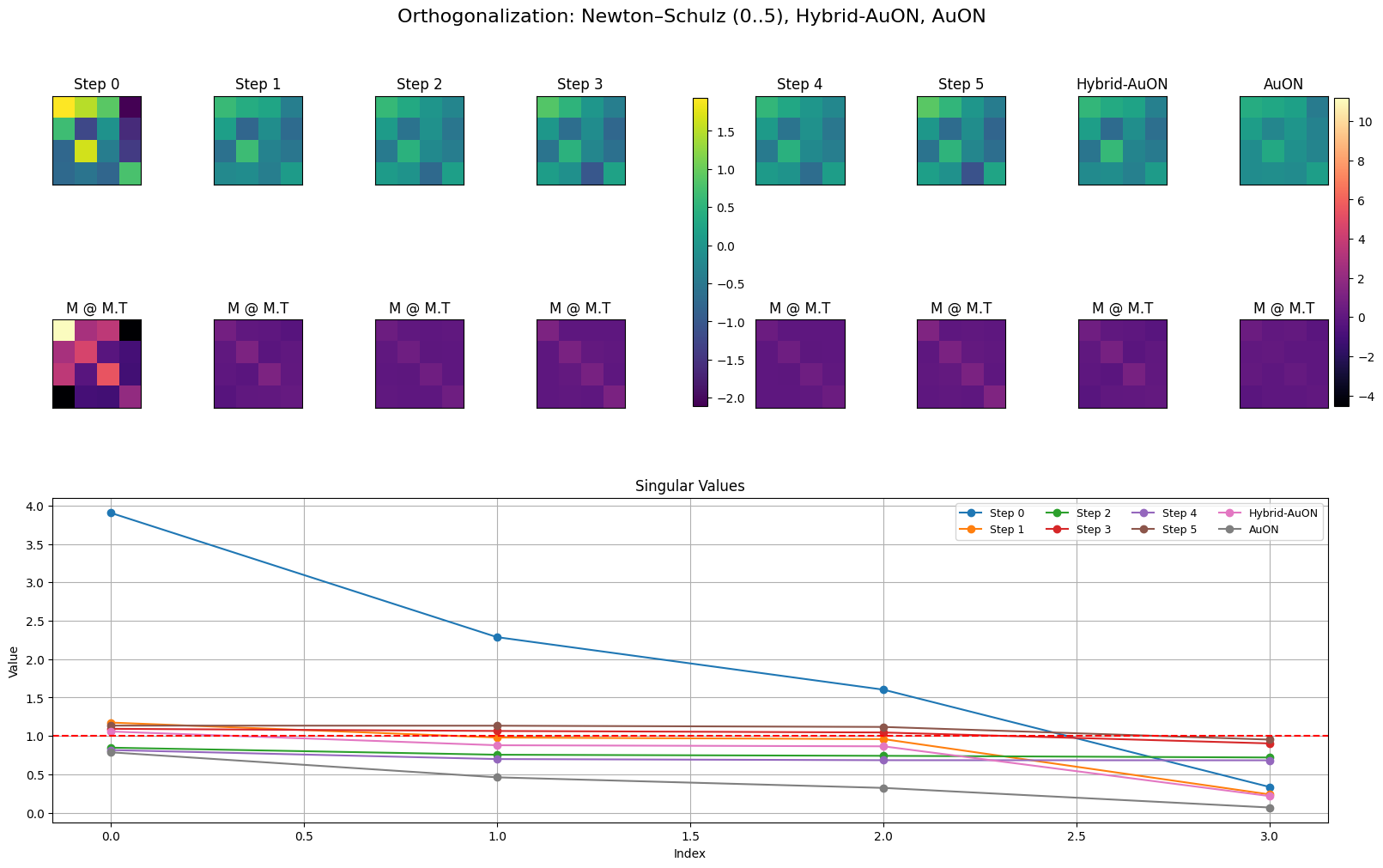}
    \caption{Visualization of the Newton--Schulz process (0.5) over 5 iterations, compared with AuON and Hybrid-AuON(NS=1). 
    The heatmaps (top) show progressive orthogonalization, with $M M^\top$ converging from a scattered structure (Step 0) to an identity-like diagonal (Step 5). 
    The singular value plot (bottom) illustrates rapid convergence toward $1.0$, confirming orthogonalization.}
    \label{fig:training_curves}
\end{figure}
\cite{tuddenham2022orthogonalisinggradientsspeedneural} proposed an approach for neural network optimization in which the gradient is first orthogonalized via singular value decomposition (SVD), followed by the application of momentum, and then the resulting momentum term is used as the update. They refer to this method as Orthogonal-SGDM. In their experiments, they observed that even in the best-performing configuration, Orthogonal-SGDM was outperformed by a well-tuned standard SGD with momentum. This is because applying momentum after orthogonalization damages the momentum mechanism: orthogonalizing gradients before accumulation prevents momentum from effectively reducing variance and maintaining beneficial directional information. Moreover,  orthogonality erases singular-value magnitudes and over-constrains the step, collapsing its singular-value structure to an isometry. In effect, the update becomes a spectral-norm-constrained move that discards useful magnitude information, wiping out correlations between update directions. Making all updates to unit length may also increase harmful alignment effects. 
Recent advances, such as Muon \cite{jordan2024muon}, improve efficiency and performance by approximating an orthogonal matrix using Newton--Schulz iterations rather than a full orthogonal matrix using SVD, and by reordering the momentum update to occur before orthogonalization. This makes the Muon computational efficient in GPU, reduces the wall-clock time, and gives much higher throughput, though Muon has cubic complexity, is 2 x faster than Adam and AdamW, but still suffers from exploding attention logits.
We find out that, by
 bounding the momentum updates under a spectral-norm trust region can preserve
 the directional information without requiring explicit orthogonalization

In this paper, we focus on developing an alternative approach to bound updates with high condition numbers under a unit-norm constraint, \emph{without explicit Orthogonalization}. Our goal is to achieve strong performance with without compromising efficiency or speed. Empirically, we find that normalization followed by a hyperbolic-cosine scaling transformation can bound the updates with high condition numbers under the spectral-norm trust region and uses an automatic ”emergency brake” to be significantly more resistant to exploding attention logits than Muon, especially in heavy‑tailed regimes.

\section{Preliminaries}

\subsection{orthogonalization}

By orthogonalizing an update matrix 
$G \in \mathbb{R}^{m \times n}$ 
with singular value decomposition 
\[
G = U \Sigma V^{\top},
\]
The update is replaced by its orthogonal polar factor
\[
Q := U V^{\top}.
\]

This satisfies
\[
Q^{\top}Q = I_n \quad \text{when } m \geq n,
\qquad
QQ^{\top} = I_m \quad \text{when } m \leq n,
\]
thereby discarding the scaling information carried by the singular values 
$\Sigma$ while preserving the directional subspaces encoded by the left and right singular vectors $U$ and $V$.

In this sense, the resulting update behaves as unit-norm in the spectral domain-
\[
\|Q\|_{2} = 1
\]
with a flat singular spectrum-emphasizing the geometric structure of the optimization landscape rather than the raw gradient magnitudes.

Intuitively, this equalizes per-direction gain: directions that originally had small singular values ("rare directions") are relatively amplified while dominant directions are relatively attenuated, promoting exploration across all relevant directions and mitigating the dominance of a few high-energy modes. 

In practice, orientation and step size can be decoupled by using 
\[
\alpha Q,
\qquad 
\alpha = \frac{\|G\|_{F}}{\sqrt{\mathrm{rank}(G)}},
\]
so that scale is controlled externally while orthogonalization enforces well-conditioned, balanced updates-yielding more stable and equitable learning dynamics compared to conventional gradient-descent steps.

\subsection{Semi-Orthogonalization}
Given $G \in \mathbb{R}^{m \times n}$ with singular value decomposition
\[
G = U \Sigma V^{\top},
\]
strict orthogonalization replaces $G$ by its polar/Stiefel projection
\[
Q := U V^{\top},
\]
collapsing the singular spectrum to $\sigma_i(Q) = 1$ on the update subspace and making $Q$ an isometry with
\[
\|Q\|_{2} = 1, 
\qquad 
Q^{\top} Q = I_n \; \; (\text{or } \; QQ^{\top} = I_m),
\]
i.e., the Frobenius-nearest semi-orthogonal matrix that removes the amplitude information in $\Sigma$. \cite{article2}

In the singular basis,
\[
G^{\top} G = V \Sigma^{2} V^{\top}
\]
becomes
\[
Q^{\top} Q = I,
\qquad
QQ^{\top} = U I U^{\top} = \Pi_{\mathrm{col}(G)},
\]
turning the step into a spectral-norm-bounded move that can discard curvature-aligned anisotropy.  

Geometrically, for Muon's RMS-to-RMS operator norm, we have
\[
Q \in \arg\max_{\|X\|_{\mathrm{RMS}\to\mathrm{RMS}} \leq 1} \langle X, G \rangle,
\]
which is the linear minimization oracle (LMO) of a conditional-gradient step. Hence, the singular values are flattened; by contrast, on the standard spectral-norm ball, the LMO yields the rank-1 solution $u_{1} v_{1}^{\top}$.\cite{lee2021_vonneumann}

To avoid overconstraint, semi-orthogonal schemes such as Muon orthogonalize only the momentum $M_t$ to
\[
Q_t = \mathrm{polar}(M_t),
\]
and decouple scale via an RMS-to-RMS factor $\alpha$, giving
\[
W_{t+1} = (1 - \eta_t \lambda) W_t + \eta_t \, \alpha \, Q_t.
\]
In practice, $Q_t Q_t^{\top}$ is computed efficiently via a low-order Newton Schulz iteration, and $\alpha$ is chosen to match update RMS across shapes, enabling stability and learning-rate transfer. Semi-orthogonalization stabilizes training by bounding spectral energy and equalizing directional gains, preventing overshoot along sharp curvature, reducing oscillations, and enabling larger learning rates by decoupling orientation from scale\cite{liu2025muonscalablellmtraining}

\subsection{Orthogonalized Momentum as a Spectral Trust-Region Method}

Recent advances demonstrate that orthogonalized momentum in deep learning optimizers, particularly the Muon optimizer, admits a principled interpretation as the solution to a non-Euclidean trust-region subproblem under the spectral norm constraint~\cite{kovalev2025understandinggradientorthogonalizationdeep}. The core update rule can be formulated as
\[
X_{k+1} = X_k - \eta O_k,
\quad 
O_k = \mathrm{Orth}\!\left(\nabla F(X_k)\right),
\]
where $\mathrm{Orth}(\cdot)$ denotes the SVD-based orthogonalization operator that computes
\[
M = U \Sigma V^\top 
\quad \Longrightarrow \quad 
\mathrm{Orth}(M) = U V^\top,
\]
yielding the steepest descent direction under the spectral norm metric.

\textbf{Momentum Integration}
The momentum component follows the exponential moving average
\[
m_{k+1} = (1-\alpha)m_k + \alpha \, g(x_k;\xi_k),
\]
where $g(x_k;\xi_k)$ represents an unbiased stochastic gradient estimate. The orthogonalized update then solves the trust-region subproblem
\[
x_{k+1} 
= \arg\min_{x} 
\Big\{ \langle \mathrm{Orth}(m_{k+1}), \, x \rangle 
: \, \|x - x_k\|_2 \leq \eta \Big\}.
\]
This formulation explicitly constrains parameter updates within a trust region while ensuring the search direction maintains unit spectral norm.

\textbf{Theoretical Advantages}
The orthogonalization-first approach provides superior variance reduction compared to alternative momentum--orthogonalization orderings. By applying orthogonalization to the momentum vector before the parameter update, the method preserves the accumulated directional information while eliminating scale-dependent instabilities. This design choice demonstrates both theoretical guarantees for convergence under non-convex objectives and empirical improvements in training stability across diverse architectures \cite{liu2025muonscalablellmtraining}.

\section{Methods}
We hypothesize that forcing all update directions to unit length can be problematic, as not all directions contribute equally to optimization progress, some may be harmful (having a negative impact), or irrelevant to loss reduction.At first glance, it seems impossible: how can one suppress "harmful" directions without knowing what they are in advance? If one never explicitly decorrelates or identifies bad modes, how does it preferentially suppress them?

The answer is that AuON exploits a statistical property of bad directions: they tend to be spiky.

Most directions contribute equitably to the loss landscape; harmful or noisy directions are statistically rare and often appear as outliers—components that are much larger than average. When a harmful direction emerges, it appears as a spike or heavy-tailed component in the update matrix

Harmful directions produce spiky gradients—they are discontinuous, scattered, and their components are extremely non-uniform. Normal, beneficial updates are smoother and have more balanced component magnitudes.Not all spiky updates are harmful and not all rare direction are spiky but spiky directions carry the potentially highest risk of maximum harm.Spiky updates are high-variance "gambles." If the spike hits a sensitive parameter, it causes massive damage (high loss).

Our goal is to develop a method that suppress the harmful directions(spiky-gradient updates) or alignments and preserves the beneficial properties of orthogonalization while selectively scaling directions under unit-norm: one solution is to decrease the scales of rare update directions relative to dominant ones and keep them all under a unit-norm trust region, meaning prioritizing directions with favorable conditioning that correspond to well-conditioned subspaces of the loss landscape under a spectral-norm trust-region.To do that we can apply a temperature-scaled softmax update matrix, followed by L2-renormalization, to bound the step under a trust-region. But computing softmax may be problematic as it introduces computational bottlenecks and it can also dampens useful directions as a side effect of dampening, which is crucial for effective learning.
We empirically find that the  normalization with hyperbolic functions (cosh ) helps us bound the momentum updates under spectral-norm trust-region \cite{kovalev2025understandinggradientorthogonalizationdeep} and it is a clever statistical mechanism (the cosh-RMS scaling) that automatically dampens harmful directions as a side effect of dampening all spiky components. As a result we get "emergency brake” property which can handle 
exploding attention logits than Muon, especially in heavy-tailed regimes."Emergency Brake" is an emergent property of the hyperbolic cosine (cosh) normalization step.
\begin{figure}[!htb]
    \vspace*{\fill}
    \centering
    \includegraphics[width=0.5\linewidth]{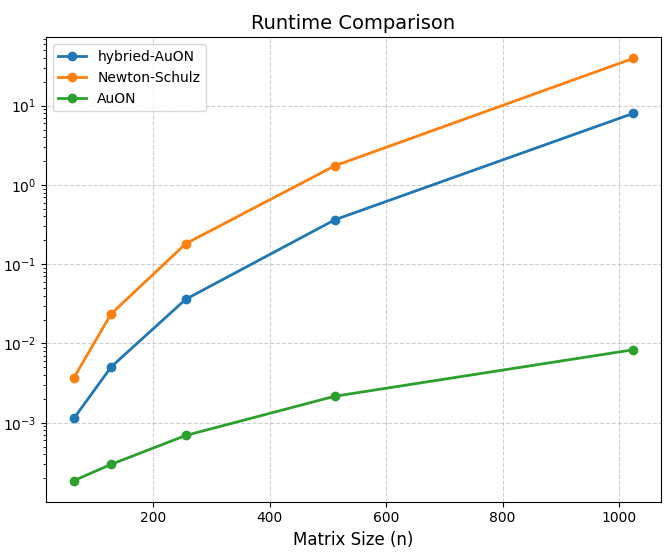}
   \caption{comparison of computation efficiency of different methods on (nxn) random matrices, where Hybrid-Auon(NS=1) }
    \label{fig:training_curves2}
    \vspace*{\fill}
\end{figure}

Which makes the training  are more stable and equitable learning dynamics compared to conventional gradient-descent steps, and stabilizes training by bounding spectral  energy into a unit vector and equalizing directional gains, preventing overshoot along sharp curvature, reducing oscillations, and
enabling larger learning rates by decoupling orientation from scale\cite{Peletier_2023}

\subsection{Nonlinear reshaping via hyperbolic cosine RMS scaling}
Our main goal is to keep all the updated directions under the unit spectral norm and remove the harmful directions. We empirically find out that the updated matrix divided by the scale factor of the RMS magnitude of cosh() helps us bound the dominant update directions that have a high condition number under unit spectral norm, and helps us to preserve the orthogonal-like properties. By doing this, we remove the harmful alignment and stabilize the updates and equitable learning dynamics. The overall equation is

\[ X = \frac{G}{\|G\| + 10^{-7}} \] \[ \text{update} = X \] \[ x = \cosh(\text{update}) \] \[ \text{rms} = \sqrt{\frac{1}{N} \sum_{i=1}^{N} x_i^2} \] \[ G = \frac{\text{update}}{\text{rms} + 10^{-8}} \]

where

\[
\cosh(z) = \frac{e^z + e^{-z}}{2}
\]

For large values of \( |z| \), \(\cosh(z)\) grows exponentially, while for small values of \(z\),
\[
\cosh(z) \approx 1 + \frac{z^2}{2}.
\]

Thus, \(\cosh\) magnifies meaningful deviations while remaining symmetric and smooth.

This encourages a spread of activations (diversity) without enforcing strict  orthogo- \indent nality \cite{Peletier_2023}, and X is the updated momentum vector direction 

\noindent\textbf{Effect of the Hyperbolic Cosine RMS Magnitude.}\\
Define \(\mathrm{rms} := \|\cosh(\mathrm{update})\|_{F}/\sqrt{N}\), where \(\cosh\) is applied \emph{only} to compute a global, tail-sensitive scale \cite{article}. Because \(\cosh\) is even and rapidly increasing in \(|x|\), heavy tails inflate \(\mathrm{rms}\), which reduces the overall step size when forming \(U := \mathrm{update}/(\mathrm{rms}+10^{-8})\). Crucially, \(\cosh\) is not applied to the propagated vector: \(U\) is a uniform rescaling of \(\mathrm{update}\), so the signs and all relative component ratios of \(\mathrm{update}\) are preserved in \(U\). This yields scale invariance with tail-aware damping, without introducing per-coordinate reweighting in the final update. As a result, as soon as attention logits start to explode (creating gradient spikes), AuON automatically suppresses the update to near-zero. It effectively freezes the unstable layer rather than pushing it further.it is tail-aware. It uses the cosh function to detect when the optimization landscape is becoming unstable (spiky gradients) and reacts by shrinking the step size

\medskip
\noindent\textbf{Intuitive interpretation.} 
The procedure can be seen as a three-step normalization: 
(1) fix the raw step's size, 
(2) evaluate how concentrated or spiky it is using the cosh statistic, 
(3) Shrink the entire step proportionally if it is highly spiky. 
Importantly, the internal proportions and direction of the update remain unchanged.

\noindent\textbf{Spectral Trust-Region Normalization.}

\noindent\textbf{Exact orthogonality.}
A matrix \(W \in \mathbb{R}^{m \times n}\) is orthogonal (semi-orthogonal if \(m \neq n\)) when
\[
W^\top W = I_n \quad \text{or} \quad W W^\top = I_m ,
\]
which preserves Euclidean inner products and hence norms and angles exactly.

\medskip
\noindent\textbf{Method (equations).}
Let \(G \in \mathbb{R}^{m \times n}\) be a gradient/update and \(N := mn\). Define
\[
\mathrm{update} := \frac{G}{\|G\|_{F} + 10^{-7}}, 
\qquad
\mathrm{rms} := \frac{\big\| \cosh(\mathrm{update}) \big\|_{F}}{\sqrt{N}},
\qquad
U := \frac{\mathrm{update}}{\mathrm{rms} + 10^{-8}}.
\]
Equivalently, \(U = X/(r+\varepsilon)\) with \(X=\mathrm{update}\), \(r=\mathrm{rms}\), \(\varepsilon=10^{-8}\).

\medskip
\noindent\textbf{Immediate implications.}
\begin{itemize}
\item \emph{Scale invariance.} For any \(c>0\), replacing \(G\) by \(cG\) leaves \(\mathrm{update}\) (and thus \(U\)) unchanged up to \(\varepsilon\)-terms.
\item \emph{Tail-aware global scaling.} Heavy tails inflate \(\mathrm{rms}\) via \(\cosh\), reducing the global magnitude of \(U\) when the update is spiky.
\end{itemize}

\medskip
\noindent\textbf{Discussion.}  
Unlike semi-orthogonal updates, AuON does not explicitly decorrelate directions. Instead, it contracts the update into a spectral-norm trust region while preserving directional ratios. This yields a \emph{correlation-preserving normalization} that dampens spiky updates but avoids the quadratic cost of orthogonalization.

\medskip
\noindent\textbf{Norms and "balanced sphere."}
This construction does not enforce unit RMS for \(U\). Indeed,
\begin{align}
\|\mathrm{update}\|_{F} \approx 1,\qquad\\
\|U\|_{F} = \frac{\|\mathrm{update}\|_{F}}{\mathrm{rms} + 10^{-8}} \approx \frac{1}{\mathrm{rms} + 10^{-8}},\\
\qquad\mathrm{RMS}(U) = \frac{\|U\|_{F}}{\sqrt{N}} \approx \frac{1}{\sqrt{N}\,(\mathrm{rms}+10^{-8})}\ \end{align}
Thus, there is no unit-L2 or unit-RMS constraint on \(U\); the overall step length decreases as the tail-sensitive scalar \(\mathrm{rms}\) increases.

\medskip
\noindent\textbf{Relation to near semi-orthogonality.}
Let 
\[
M := U^\top U \in \mathbb{R}^{n \times n}.
\]
By the Frobenius-trace identity,
\[
\operatorname{tr}(M) = \|U\|_F^2,
\]
and
\[
\alpha := \frac{1}{n} \operatorname{tr}(M),
\]
which equals the average column $\ell_2$ norm squared.  

However, under the mapping above, $\operatorname{tr}(M)$ is determined by $\mathrm{rms}$ and is not generally $N = mn$ unless an extra unit-RMS rescale is applied to $U$.  

Off-diagonal correlations 
\[
M_{ij} = \langle U_{:i}, U_{:j} \rangle
\]
are not explicitly zeroed by this mapping, so it promotes scale invariance and approximate isotropy rather than exact semi-orthogonality.
\begin{itemize}
\item \emph{Cross-correlations.} Off-diagonals of \(M\) are scaled copies of those in \(\mathrm{update}^\top \mathrm{update}\); they are not explicitly suppressed.
\item \emph{Isotropy.} This mapping alone does not drive \(M\) toward \(\alpha I_n\). Achieving near semi-orthogonality typically requires an additional correlation-reducing step (e.g., per-column RMS normalization, light whitening, or a spectral penalty \(\|U^\top U - \alpha I\|_F^2\) with \(\alpha=\tfrac{1}{n}\operatorname{tr}(U^\top U)\)). See Appendix A for more information
\end{itemize}

\noindent\textbf{Practical implication.}\\
The update is scale-invariant and tail-aware: heavy tails trigger stronger global shrinkage via \(\mathrm{rms}\), helping prevent blow-ups while preserving the direction and internal proportions of the step. When approximate isotropy or near semi-orthogonality is desired, pair this normalization with a lightweight correlation-suppressing operation.

\subsection{Hybrid Approach}

While AuON provides stability through spectral trust-region normalization, it does not 
reduce inter-directional correlations. In contrast, Muon applies several iterations of the 
Newton-Schulz process to approximate semi-orthogonalization, which is effective but 
computationally expensive (\(O(n^2)\) per iteration).

We therefore introduce a \emph{hybrid} scheme: apply a single Newton-Schulz iteration(Auon+) or 3(AuON+3) or 5(AuON+5) Newton-Schulz iterations 
to partially decorrelate the update, followed by AuON rescaling to enforce spectral 
contraction.

Let \(X\) denote the update matrix. A single Newton-Schulz step produces
\[
A = XX^\top, 
\qquad 
B = bA + cA^2, 
\qquad 
X \leftarrow aX + BX ,
\]
for scalars \(a,b,c\) ensuring contraction.  
Then apply AuON normalization:
\[
\mathrm{update} := \frac{X}{\|X\|_F + \epsilon_0}, 
\qquad
r := \frac{\|\cosh(\mathrm{update})\|_F}{\sqrt{N}}, 
\qquad
U := \frac{\mathrm{update}}{r + \epsilon}.
\]

\noindent\textbf{Properties.}
\begin{itemize}
    \item \emph{Partial decorrelation.} The single Newton-Schulz step reduces but does 
    not eliminate correlations between directions.
    \item \emph{Spectral contraction.} The AuON scaling guarantees a strict spectral 
    trust-region bound, stabilizing the update.
    \item \emph{Efficiency.} Compared to Muon's multi-step Newton-Schulz iterations 
    (5-10 steps), AuON requires only one iteration, reducing overhead while 
    retaining some correlation control.
\end{itemize}

\medskip
\noindent\textbf{Discussion.}  
Hybrid-AuON is not a full semi-orthogonalization method. It preserves part of the 
correlation structure, but guarantees bounded, tail-aware updates. Its purpose is to offer 
a middle ground: more stable than plain AuON, but cheaper than Muon.

\subsection{Temperature–Scaled AuON: Sensitivity to Tensor Size and Heavy Tails}
\label{subsec:temp-auon}
In temperature-scaled AuON, the \emph{cosh input scale} $\gamma$ (parameterized by
the exponent $\alpha$) and the \emph{RMS exponent} $\beta$ are the primary
hyperparameters determining how early and how aggressively the brake engages on
heavy-tailed updates.  
\[
\tilde G = \frac{G}{\lVert G\rVert_F},
\]
then applies a temperature
\[
\gamma = N^{\alpha}, \qquad N = \mathrm{numel}(\tilde G),
\]
to obtain
\[
g_{\text{scaled}} = \gamma\,\tilde G.
\]
The brake statistic is computed as
\[
r = \Bigl(\mathbb{E}\big[\cosh^2\!\big(g_{\text{scaled}}\big)\big]\Bigr)^{\beta/2},
\]
and the update takes the form
\[
\Delta\theta = -\,\mathrm{lr}\,\frac{G}{r + \varepsilon}.
\]

In this AuON variant, the temperature is set to $\gamma = N^{0.48}$, where $N$ is the
number of elements in the update tensor, and the hyperbolic--cosine RMS statistic
is raised to the power $3.5/2$. These two modifications make the brake extremely
sensitive to both tensor width and the presence of heavy-tailed updates.

The update pipeline is:
\begin{equation}
g \;\xrightarrow{\text{momentum}}\;
g_{\text{hat}}
= \frac{g}{\|g\| + \varepsilon}
\;\xrightarrow{\text{temperature}}\;
g_{\text{scaled}} = N^{\alpha} g_{\text{hat}},\qquad \alpha = 0.48,
\end{equation}
\begin{equation}
s = \mathbb{E}\!\left[\cosh^2\!\left(g_{\text{scaled}}\right)\right],
\qquad
r = s^{\,3.5/2},
\qquad
\Delta\theta = -\,\text{lr}\;
\frac{g}{\,r + 10^{-8}}.
\end{equation}

For small arguments, $\cosh(x) \approx 1 + x^{2}/2$, giving
\(
s \approx 1 + \mathbb{E}[g_{\text{scaled}}^{2}],
\)
but since $g_{\text{scaled}}$ has variance proportional to $N^{2\alpha}$, even
The quadratic term grows with tensor size.  
For rare large entries, $\cosh(x) \sim \tfrac{1}{2} e^{|x|}$, so a tiny number of
outliers with $|x|\gg 1$ dominate $s$ exponentially.  
Raising $s$ to the exponent $3.5/2$ further \emph{super-linearizes} this effect:
If the tail magnitude doubles, the brake factor $r$ grows by more than a factor of two,
causing the step norm to collapse sharply.

The temperature parameter $\alpha$ controls cross-layer behavior.
Smaller $\alpha$ slows the growth of $N^{\alpha}$, keeping
$g_{\text{scaled}}$ mostly in the mild regime ($|x|\lesssim 1$) even in wide layers.
This yields more uniform braking across tensor shapes and makes AuON resemble a soft,
global trust-region scaling.  
Larger $\alpha$ amplifies $N^{\alpha}$, causing $r$ to explode in wide layers,
which dramatically reduces their effective learning rate and strongly suppresses any
bursty or sharp gradients.

\subsection{Spectral Trust-Region Guarantee and Reduction of Correlation Energy in AuON}

Let $G \in \mathbb{R}^{m \times n}$ be an update matrix, and define the normalized update
\[
\widetilde{G} = \frac{G}{\|G\|_F + \varepsilon_0}, 
\quad N = mn, 
\quad \varepsilon_0 > 0.
\]
Define
\[
r = \frac{1}{N}\sum_{i,j} \cosh^2(\widetilde{G}_{ij}), 
\qquad U = \frac{\widetilde{G}}{r+\varepsilon}, 
\quad \varepsilon > 0.
\]

The AuON transformation $G \mapsto U$ guarantees the following:

\paragraph{(1) Spectral Trust-Region Bound.}
For every update $U$ produced by AuON, the spectral norm is strictly contractive:
\[
\|U\|_2 \;\leq\; \frac{1}{r+\varepsilon} \;\leq\; \frac{1}{1+\tfrac{1}{N}+\varepsilon} \;<\; 1.
\]

\emph{Proof.} By construction,
\[
U = \frac{\widetilde{G}}{r+\varepsilon},
\qquad
\|U\|_2 = \frac{\|\widetilde{G}\|_2}{r+\varepsilon}
\;\leq\; \frac{\|\widetilde{G}\|_F}{r+\varepsilon}.
\]
Since $\widetilde{G} = G/(\|G\|_F+\varepsilon_0)$, we have the \emph{squared} Frobenius norm
\[
\|\widetilde{G}\|_F^2
= \frac{\|G\|_F^2}{(\|G\|_F+\varepsilon_0)^2}
\;\approx\; 1
\]
(up to $\varepsilon_0$-terms). Hence $\|\widetilde{G}\|_F \approx 1$, so
$\|U\|_2 \leq 1/(r+\varepsilon)$.

Now using $\cosh(x) \geq 1 + \tfrac{x^2}{2}$, we obtain
\[
r^2 = \tfrac{1}{N}\sum_{i,j}\cosh^2(\widetilde{G}_{ij})
\;\;\geq\;\; 1 + \tfrac{1}{N}\sum_{i,j}\widetilde{G}_{ij}^2
= 1+\tfrac{1}{N}\|\widetilde{G}\|_F^2
\;\;\approx\;\; 1 + \tfrac{1}{N}.
\]
Thus $r \geq 1+1/N$, which is the \emph{tightest possible lower bound}
without additional assumptions on the distribution of entries in $G$. 
Substituting yields the spectral guarantee. \hfill $\square$

\paragraph{(2) Tail Sensitivity and Spike Suppression.}
If some entry of $\widetilde{G}$ satisfies $|\widetilde{G}_{ij}| = a$, then
\[
r \;\geq\; \tfrac{\cosh(a)}{N},
\qquad
\|U\|_2 \;\leq\; \tfrac{N}{\cosh(a)}.
\]
Hence, large spikes are suppressed at an exponential rate:
\[
\|U\|_2 \;\lesssim\; \tfrac{2}{N}e^{-a}, 
\qquad (a\to\infty).
\]

\emph{Proof.} A single large coordinate contributes at least $\cosh^2(a)/N$ to $r^2$, so $r \geq \cosh(a)/N$. Substituting into $\|U\|_2 \leq 1/(r+\varepsilon)$ yields the result. Since $\cosh(a)\sim \tfrac{1}{2}e^a$ as $a\to\infty$, exponential decay follows. \hfill $\square$

\paragraph{(3) Reduction of Correlation Energy.}
Let $M = U^\top U$. Then
\[
M = \frac{\widetilde{G}^\top \widetilde{G}}{(r+\varepsilon)^2}.
\]
Consequently,
\[
\operatorname{tr}(M) \;=\; \frac{\|\widetilde{G}\|_F^2}{(r+\varepsilon)^2} \;\approx\; \frac{1}{(r+\varepsilon)^2},
\]
and
\[
\Bigl\|M - \tfrac{1}{n}\operatorname{tr}(M)I\Bigr\|_F
= \frac{\|\widetilde{G}^\top \widetilde{G} - \tfrac{1}{n}\operatorname{tr}(\widetilde{G}^\top \widetilde{G})I\|_F}{(r+\varepsilon)^2}.
\]
Thus both the total spectral energy and the isotropy residual contract by the deterministic factor $(r+\varepsilon)^{-2}$. \hfill $\square$

\paragraph{Interpretation.}
Unlike previous semi-orthogonal updates, AuON achieves in a \emph{single linear-time pass}:
(i) strict spectral trust-region safety ($\|U\|_2 < 1$),
(ii) exponential damping of spiky components, and
(iii) uniform contraction of correlation energy toward near isotropy.  
These properties together provide a lightweight but principled alternative to iterative orthogonalization.

\subsection{Emergent ``Emergency Brake'' Behavior in AuON via Hyperbolic-Cosine Normalization}
 A critical failure mode in large-scale Transformer training is the phenomenon of “exploding at
tention logits.” As weight norms grow unbounded, the query-key dot products diverge, causing
 Softmax saturation. This results in gradient matrices that are geometrically sparse and heavy-tailed (high kurtosis).

 \textbf{AuON } introduces a hyperbolic-cosine-based–based normalization that acts as an automatic “emergency brake” when gradients become unstable, for example under exploding attention heads. Unlike Muon, which normalizes to unit norm regardless of the gradient distribution, AuON computes a scalar factor
\subsubsection{Cosh-Based Normalization as an Instability Detector: The Hyperbolic Sensor}

We demonstrate that AuON possesses an inherent self-regulating mechanism—the \textit{Emergency Brake}—which is an emergent property of the hyperbolic cosine scaling. Unlike semi-orthogonal projections that enforce unit energy updates regardless of landscape curvature, AuON utilizes the $\cosh(\cdot)$ function as a non-linear stability sensor.

The core stability mechanism lies in the construction of the normalization scalar $r$. Recall the temperature-scaled definition:
\begin{equation}
    r = \sqrt{\frac{1}{N} \sum_{i,j} \cosh^2\left(\gamma \cdot \tilde{G}_{ij}\right)} \quad \text{where } \gamma = \sqrt{N}
\end{equation}
The function $f(x) = \cosh(x)$ exhibits two distinct asymptotic behaviors depending on the magnitude of the input gradient:

\begin{enumerate}
    \item \textbf{The Safe Zone (Quadratic Regime):} For well-behaved gradients where values are near zero, $\cosh(x) \approx 1 + x^2/2$. In this regime, $r$ behaves similarly to a standard $L_2$ norm, facilitating efficient optimization.
    \item \textbf{The Danger Zone (Exponential Regime):} For large outliers (spikes) caused by exploding logits, $\cosh(x) \approx \frac{1}{2}e^{|x|}$.
\end{enumerate}

Consider a scenario where the gradient matrix develops an instability spike of magnitude $\xi \gg 1$. The normalization factor $r$ grows exponentially:
\begin{equation}
    r \propto e^{|\xi|}
\end{equation}
Consequently, the final update norm $||U||_2 = ||\tilde{G}||_2 / r$ is suppressed:
\begin{equation}
    ||U||_2 \propto \frac{\xi}{e^{|\xi|}} \xrightarrow{\xi \to \infty} 0
\end{equation}
This proves that AuON automatically throttles the update size towards zero as instability increases, effectively freezing the weights of the unstable layer to allow for energy dissipation.
In effect, cosh acts as a nonlinear sensor that is nearly quadratic in the “safe” regime
but becomes extremely sensitive to rare, large-magnitude components.

\

\subsubsection{Sequence of Brake Activation Under Exploding Attention}

When an attention head becomes unstable, attention logits can grow excessively large, driving the softmax into saturation and producing a gradient matrix $\tilde G$ that is sparse and spiky: most entries remain near zero, while a small subset corresponding to unstable tokens attain very high magnitudes (e.g., on the order of $10$ after pre-normalization).  

In conventional $\ell_2$-based normalization, a single spike of magnitude $10$ contributes only $10^2 = 100$ to the norm sum over millions of parameters---negligible in the global scale and therefore ineffective for damping such instabilities.

Under AuON, the same spike passes through $\cosh$: for $\tilde G_{ij} = 10$, $\cosh(10)$ is on the order of $10^4$, and AuON further squares this, yielding $\cosh^2(10) \approx 10^8$. Even after averaging over all $N$ parameters, this single outlier term dominates the computation of $r$, causing the normalization factor to grow dramatically relative to its normal value.

\subsubsection{Global Suppression as the ``Emergency Brake''}

The final AuON update is defined as
\begin{equation}
U = \frac{\tilde G}{r}.
\end{equation}
Once a spike drives $r$ to a large value, the entire update matrix $U$ is proportionally downscaled.  
The offending spiky entries are strongly suppressed, and the otherwise normal entries are simultaneously reduced toward zero, effectively freezing that layer’s update for the current step.  
This global suppression acts as an \textbf{automatic emergency brake}: it activates only when the gradient field contains extreme outliers yet leaves normal training dynamics largely unaffected when all entries of $\tilde G$ lie within the quasi-quadratic region of $\cosh$.

\subsubsection{Empirical Verification at Production Scale (d=4096)}

To stress-test the “Emergency Brake” mechanism at a realistic scale, we constructed
 a Transformer layer with model dimension d = 4096 and an MLP expansion factor of
 4(Dmlp = 4096), totaling approximately 100.6 million parameters. Crucially,
 all normalization layers (LayerNorm/RM SNorm) were removed and the learning rate was kept at 0.24 . This architecture creates a high-pressure optimization environment where weight accumulation leads directly
 to attention logit explosion. We compared the Newton-Schulz semi-orthogonal update
 (Muon) against AuON (with temperature scaling )over 1000 steps.

\begin{figure}[H]  
  \centering
  \includegraphics[width=0.85\textwidth]{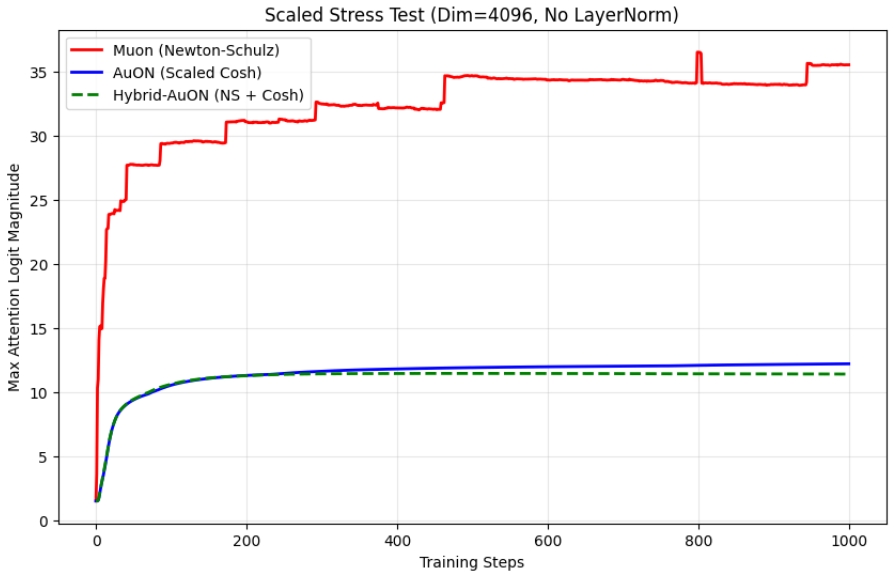}
  \caption{\textbf{The Emergency Brake in Action.} The Muon optimizer (Red) exhibits step-function explosions, driving logits to hard saturation (greater than 30). In contrast, AuON (Blue) and AuON+5(Ns=5)(Green) demonstrate the theoretically predicted logarithmic braking curve, capping logit growth at approximately 12 even after 1,000 steps of unnormalized accumulation.}
  \label{fig:brake_verification}
\end{figure}

As illustrated in Figure~\ref{fig:brake_verification}, the divergence in behavior is stark:

\begin{itemize}
    \item \textbf{Muon (No Brake):} The semi-orthogonal update ignores the growing gradient magnitude, treating the exploding logits as a strong signal. This leads to immediate saturation and chaotic quantization artifacts.
    
    \item \textbf{AuON and Hybrid-AuON(Ns=5) (Brake Engaged):} The blue and green trajectories confirm the exponential suppression derived in Eq.~(9). As logits attempt to grow, the gradient kurtosis rises, triggering the $\cosh$ term to inflate the denominator $r$. The result is a smooth, controlled saturation curve rather than a catastrophic explosion.
\end{itemize}

\subsubsection{Long-Horizon Stability: The Dissipative Property}

While short-term stress tests reveal the immediate reaction to shock, long-horizon training reveals the fundamental thermodynamic properties of the optimizer. We extended the unnormalized Transformer stress test ($d=1024$) to 20,000 steps to observe the asymptotic behavior of the logit magnitudes.

\begin{figure}[H]
  \centering
  \includegraphics[width=0.85\textwidth]{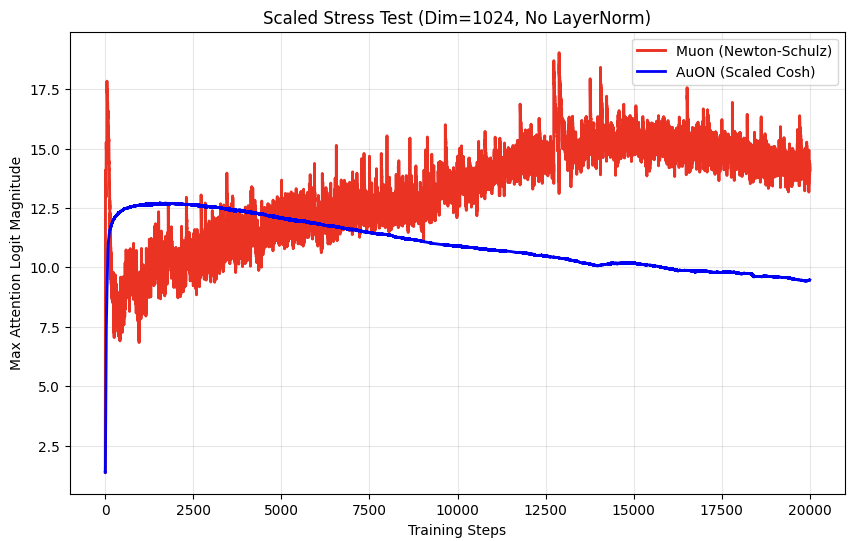}
  \caption{\textbf{Thermodynamics of Optimization (20k Steps).} The Muon optimizer (Red) exhibits high-variance chaotic behavior with a persistent upward drift, indicative of unregulated energy injection. In stark contrast, AuON (Blue) demonstrates a dissipative trajectory. After an initial rise due to the lack of normalization, the self-regulating cosh mechanism forces an inflection point at approximately step 2,500, after which the logit magnitudes steadily decay. This proves that AuON actively removes instability energy from the system over time.}
  \label{fig:long_horizon}
\end{figure}

\subsubsection{Analysis of Asymptotic Behavior}

The trajectory in Figure~\ref{fig:long_horizon} highlights a critical distinction in stability mechanics:

\begin{enumerate}
    \item \textbf{Chaotic Energy Injection (Muon):} The high variance (jaggedness) of the red curve indicates that Muon is continuously oscillating between saturation states. The spectral constraint (update norm equals 1) forces the optimizer to maintain high kinetic energy even when the loss landscape suggests caution, preventing convergence to a stable basin.
    
    \item \textbf{Controlled Energy Dissipation (AuON):} The smooth, downward trajectory of the blue curve confirms the theoretical prediction of Section 3.3. As logits grow, the gradient kurtosis increases. The temperature-scaled cosh denominator ($r$) responds exponentially, driving the effective update norm $\|U\|_2 \ll 1$. This turns the optimizer into an energy sink, draining the excess weight magnitude accumulated during the early training phase and naturally guiding the system back toward a stable regime ($\approx 9.5$) despite the absence of explicit normalization layers.
\end{enumerate}

\subsection{Empirical Validation: The Geometry of Harmful Directions}
\label{subsec:empirical2}

To validate the hypothesis that harmful optimization directions are characterized by spatial \emph{spikiness} (high kurtosis), we analyzed both real-time training dynamics and static landscape geometry.  
The results provide strong empirical confirmation that spikiness serves as a reliable geometric signature of instability.

\subsubsection{Dynamic Instability Analysis (The ``Smoking Gun'')}
\label{subsubsec:dynamic}

\begin{figure}[H]
    \centering
    \includegraphics[width=0.85\linewidth]{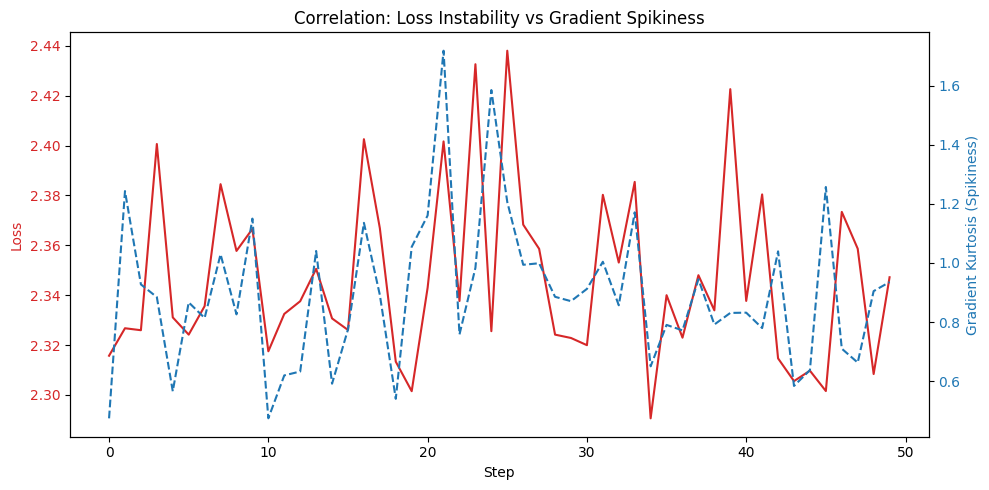}
    \caption{\textbf{Training Monitor:} Correlation between loss instability (red) and gradient kurtosis (blue).  
    \textbf{Figure 1:} Temporal correlation of instability. The training trajectory reveals that significant spikes in training loss (instability events, e.g., at steps 24, 40, and 46) are inextricably linked to simultaneous or preceding spikes in gradient kurtosis.  
    This establishes high kurtosis as a leading indicator of divergence in real-world optimization.}
    \label{fig:training_corr}
\end{figure}

Our primary evidence is derived from monitoring the \textbf{Fisher kurtosis} of gradient updates during the training of a Transformer model.  
As illustrated in Figure~\ref{fig:training_corr}, we observe a strong temporal correlation between geometric spikiness and loss instability.

The synchronization between loss spikes (red line) and gradient kurtosis (blue dashed line) confirms that when the optimizer naturally encounters ``bad directions,’’ those directions manifest as heavy-tailed, spiky distributions.  
This validates the premise that the onset of exploding attention logits creates an impulsive noise signature that can be detected and suppressed via kurtosis-sensitive normalization.

\subsubsection{Geometric Risk Analysis (The Iso-Norm Probe)}
\label{subsubsec:geometry}

\begin{figure}[H]
    \centering
    \includegraphics[width=0.85\linewidth]{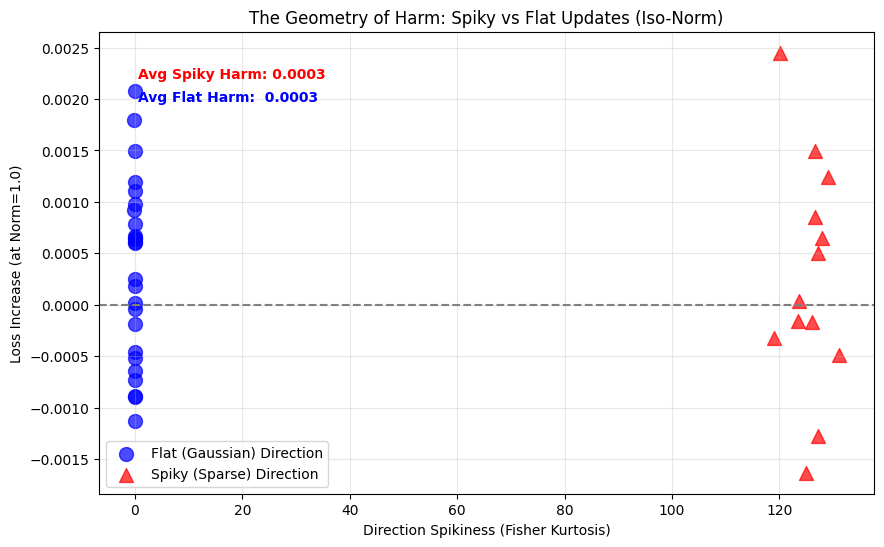}
    \caption{\textbf{Landscape Probe:} Loss increase vs.\ direction spikiness for iso-norm vectors.  
    \textbf{Figure 2:} The geometry of risk. Gaussian (flat) perturbations (blue circles) cluster tightly with low impact, while spiky perturbations (red triangles) exhibit higher variance.  
    Crucially, the most damaging directions (highest $y$-axis values) are exclusively spiky.}
    \label{fig:landscape_probe}
\end{figure}

To disentangle the effects of vector geometry from vector magnitude, we conducted a landscape probe using \textbf{iso-norm perturbations}.  
We injected random noise vectors—normalized to identical $\ell_2$ norms but differing in spatial distribution—into a converged model.  
Figure~\ref{fig:landscape_probe} presents the comparative impact on validation loss.

\paragraph{Risk Distribution.}
The analysis reveals a nuanced but critical finding regarding optimization risk:
\begin{itemize}
    \item \textbf{Average vs. Variance:} While the \emph{average harm} of random spiky directions is comparable to that of flat directions, the variance is significantly higher.  
    Flat directions represent ``safe bets’’ with predictable effects on loss.
\end{itemize}

\paragraph{Maximum Risk Confirmation.}
The single most damaging direction identified in the experiment (causing a loss increase of $+0.0025$) corresponded to a high-kurtosis vector.  
This confirms that while not all spiky directions are harmful, the most dangerous outliers in the loss landscape—those corresponding to the sharpest curvature—are statistically aligned with spiky vectors.

\subsubsection{Summary of Findings}
\label{subsubsec:summary}

Combining the dynamic evidence (Test~B) with the static probe (Test~A), we conclude that spikiness is the \emph{geometric signature of harmful optimization directions}.  
Flat (Gaussian) updates represent the dense, informative signal required for learning, while high-kurtosis updates represent high-risk instability.  

The \textbf{AuON} optimizer’s ``emergency brake’’ mechanism effectively mitigates this risk by selectively suppressing the high-variance, spiky components identified in Figure~\ref{fig:landscape_probe}, thereby preventing the divergence events visualized in Figure~\ref{fig:training_corr}.

\section{Experiments}
\subsection{ Small Language Modeling}
We evaluate our approach using a 4X L4 GPU   on the \textbf{SmolLM-Corpus} dataset\cite{benallal2024smollmcorpus} , consisting of 500k tokens. 
The underlying model is a nanoGPT\cite{Karpathy2022} with FlashAttention-2\cite{dao2023flashattention2fasterattentionbetter} rotary position embeddings (RoPE)\cite{su2023roformerenhancedtransformerrotary}, RMSNorm \cite{huang2019rmsnorm}, and SwiGLU activations \cite{shazeer2020gluvariantsimprovetransformer}. 
For the \textbf{Small configuration}, we use a hidden size of 512, 6 layers, 8 attention heads, and a feed-forward dimension of 1536. 
Training is conducted for 6000 steps with a global batch size of 128. 
We compare \textsc{AuON}, AdamW \cite{loshchilov2019decoupledweightdecayregularization}, Hybrid-AuON, and MuON under similar training conditions, with learning rates tuned separately: 
$\eta_{\text{adamw}} = 0.003$, $\eta_{\text{auon}} = 0.24$. and $\eta_{\text{muon}} = 0.01$. \cite{rosic2025muon}

\begin{table}[ht] 
\centering
\caption{Training Results on Tiny (Run 1). All optimizers are trained under identical settings.}
\label{tab:tiny_run1}
\begin{tabular}{lcccccc}
\hline
\textbf{Optimizer} & \textbf{Total Params} & \textbf{Opt. Params} & \textbf{Loss} & \textbf{Acc} & \textbf{PPL} \\
\hline

AdamW        & 40,901,120 & 40,901,120   & 0.0686 & 0.9846 & 1.07 \\
AuON         & 40,901,120 & 15,728,640 &0.0476 &  0.9897 & 1.05\\
Hybrid-AuON  & 40,901,120 & 15,728,640 & 0.0422 & 0.9908 & 1.04 \\
Muon       & 40,901,120 & 15,728,640  & 0.0375 & 0.9919 & 1.04 \\
\hline
\end{tabular}
\end{table}

\subsection{Comparison of Hybrid-AuON and Muon on FineWebEDU-10B}
\label{subsec:auon-vs-muon}

\paragraph{Setup.}
We conduct a controlled paired comparison between Temperature-scaled Hybrid-AuON and Muon on a H100 GPU with a 12-layer GPT-style decoder with hidden size $d=768$ and 6 attention heads(168 Million parameters).
Training data are streamed from the 10B-token FineWebEDU \cite{kydlicek2025finepdfs} corpus using a custom generator
that samples contiguous 8k-token sequences while respecting document boundaries via
alignment to the GPT-2 end-of-sequence token (50256).  
Each training step consumes a single 8k-token sequence for both input and target.
Validation uses the same sequence length with a fixed budget of 8 sequences (64k tokens),
evaluated every 100 steps.

Temperature-scaled variants use 
    \[
    g_{\text{scaled}} = \gamma\,\tilde{G}, \quad \text{with} \quad \gamma = N^{\alpha},
    \]
    where $N$ is the tensor size. Larger $\gamma$ amplifies outliers inside $\cosh$, increasing brake strength and $\alpha$ is 0.48.

    Vanilla AuON computes 
    \[
    r = \sqrt{\mathbb{E}[\cosh^2(\tilde{G})]},
    \]
    while temperature-scaled variants use 
    \[
    r = \big(\mathbb{E}[\cosh^2(\tilde{G})]\big)^{\beta}, \quad \beta = 3.5,
    \]
    yielding super-linear suppression that engages earlier under spiky gradients.It just change how hard and when the brake engages

\paragraph{Optimization.}
Parameters are split into two groups.
Scalar parameters, embeddings, and the LM head are optimized with Adam
($\text{lr}=0.008$, $\beta=(0.8,0.95)$, $\varepsilon=10^{-10}$, no weight decay).
Hidden matrix parameters in the transformer blocks are optimized with either:
(i)~\textbf{Hybrid-AuON}, using $\text{lr}=0.048$, momentum $=0.95$, weight decay $=0.01$,
and a 5-step Newton--Schulz orthogonalization applied only to matrix-shaped tensors; or
(ii)~\textbf{Muon}, using $\text{lr}=0.05$, momentum $=0.95$, weight decay $=0.01$,
with its standard zero-power update.  
Both share a common schedule $\mathrm{get\_lr}(\mathrm{step})$ consisting of an initial
constant phase followed by a linear decay from $1.0$ to $0.1$ over the final
$45\%$ of the 5{,}000 training iterations.  
A single RNG seed is used so that both runs start from identical initializations
and observe the same data order, ensuring a strict paired comparison.

\begin{figure}[t]
    \centering
    \includegraphics[width=\linewidth]{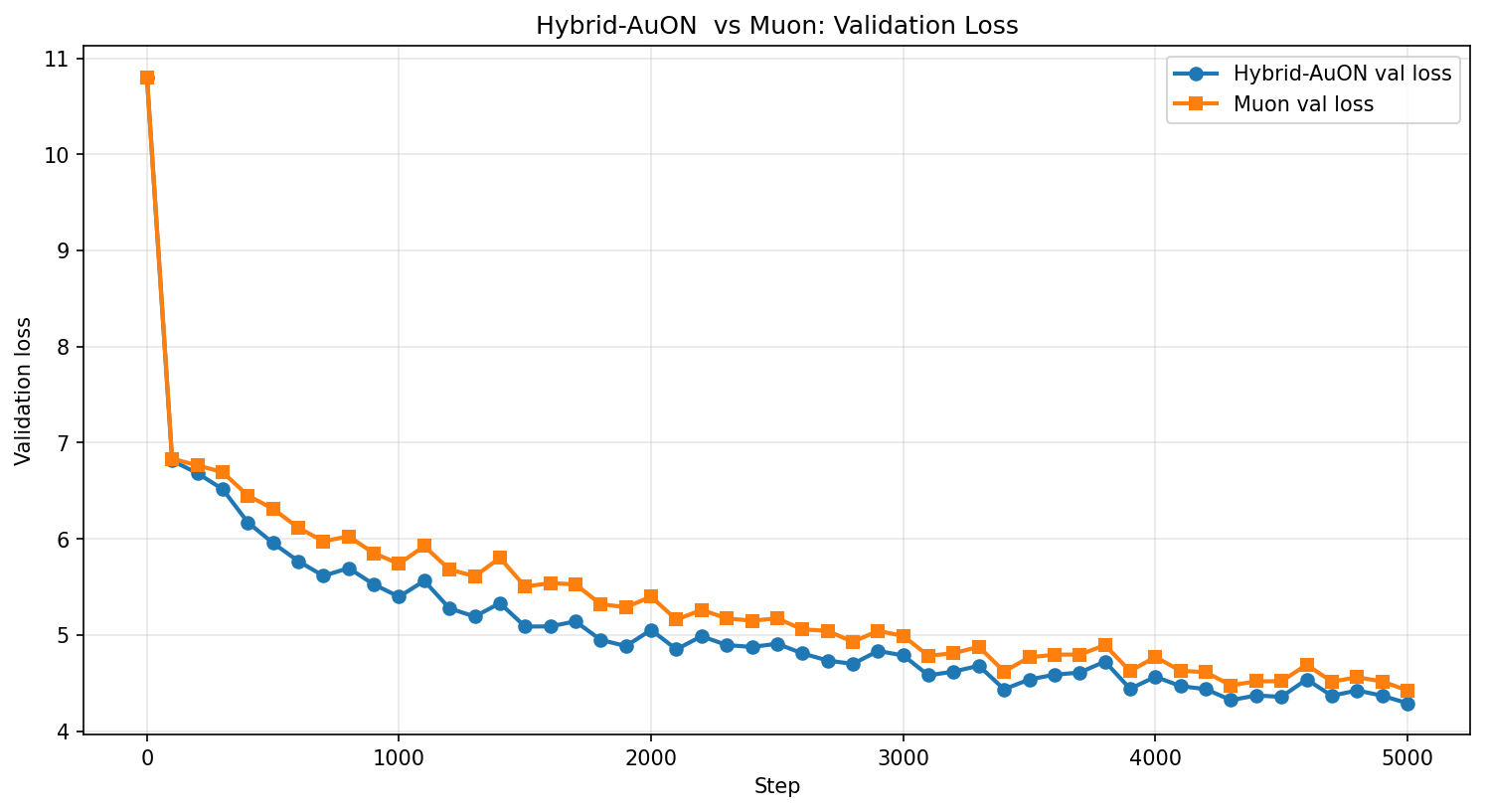}
    \caption{
    Validation loss on FineWebEDU-10B comparing Hybrid-AuON (blue) and Muon (orange).
    Both runs share identical random seeds, data streams, and Adam updates on
    non-matrix parameters. Hybrid-AuON achieves consistently lower validation loss
    throughout training, particularly in the early and mid stages.}
    \label{fig:auon-muon-val}
\end{figure}


\begin{table}[ht]
\centering
\caption{Final validation loss after 5{,}000 training steps. 
Hybrid-AuON (NS=5) achieves a lower loss than Muon.}
\label{tab:hybrid_auon_vs_muon}
\begin{tabular}{lcc}
\toprule
\textbf{Optimizer} & \textbf{Configuration} & \textbf{Final Val. Loss} \\
\midrule
Hybrid-AuON & Newton--Schulz (5 steps) & \textbf{4.2882} \\
Muon        & Zero-power orthogonal update & 4.4349 \\
\bottomrule
\end{tabular}
\end{table}

\subsection{Combined Analysis: MNIST and CIFAR-10 CNN}

We conduct a combined evaluation of \textbf{AuON}, \textbf{SGD}, and \textbf{AdamW} 
on two vision benchmarks: a simple multilayer perceptron (MLP) on MNIST and a 
convolutional neural network (CNN) on CIFAR-10. These experiments, adapted from 
Wei (2025)~\cite{wei2025muontutorial}, are designed to assess both convergence 
behavior and runtime efficiency in progressively more challenging tasks.

\paragraph{MNIST Results.}
On the MNIST classification task, AdamW achieves the lowest final training loss 
($0.0178$), as reported in Table~\ref{tab:mnist-results}. AuON attains a 
final loss of $0.0365$, substantially outperforming SGD ($0.2410$) while incurring 
only a $\sim$13\% runtime increase relative to SGD. This highlights that even in 
a simple setting, AuON improves convergence quality without straying far from 
the efficiency of standard baselines.

\begin{table}[ht]
\centering
\caption{MNIST training performance. Relative time is computed w.r.t. the fastest optimizer (SGD); lower is better.}
\label{tab:mnist-results}
\begin{tabular}{lccc}
\toprule
\textbf{Optimizer} & \textbf{Final Loss} & \textbf{Total Time (s)} & \textbf{Relative Time ($\times$)} \\
\midrule
AuON   & 0.0365 & 74.10 & 1.13$\times$ \\
SGD    & 0.2410 & 65.36 & 1.00$\times$ \\
AdamW  & 0.0178 & 69.28 & 1.06$\times$ \\
\bottomrule
\end{tabular}
\end{table}

\begin{figure}[H]
    \centering
    \includegraphics[width=0.9\textwidth]{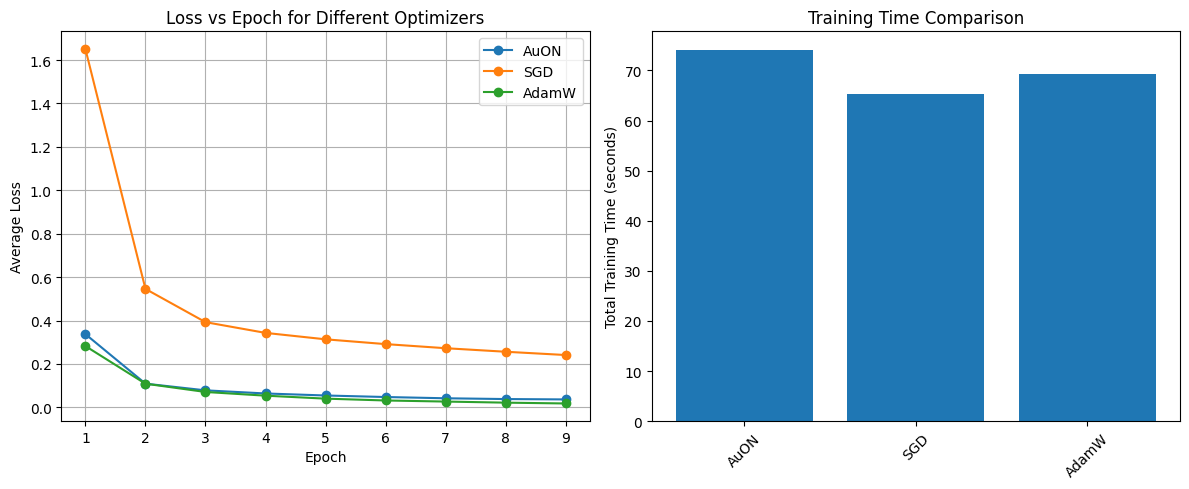}
    \caption{MNIST benchmark. Left: training loss trajectories over 9 epochs. 
    AuON converges much faster than SGD and approaches the final loss achieved 
    by AdamW. Right: total training time comparison. AuON introduces only a 
    minor overhead compared to SGD and AdamW, confirming its near-equivalent 
    efficiency.}
    \label{fig:mnist-results-png}
\end{figure}
\FloatBarrier

\paragraph{CIFAR-10 Results.}
On the more demanding CIFAR-10 benchmark, AuON demonstrates clear improvements 
in both convergence and generalization. As summarized in 
Table~\ref{tab:cifar10-results}, AuON achieves the highest test accuracy 
($74.42\%$), surpassing AdamW ($72.48\%$) and outperforming SGD by a large 
margin ($65.99\%$). Final training losses follow the same trend. Despite these 
gains, runtime remains virtually unchanged across optimizers ($\sim$88--89 seconds).

\begin{table}[ht]
\centering
\caption{CIFAR-10 CNN performance. Relative time is computed w.r.t. the fastest optimizer (SGD); lower is better.}
\label{tab:cifar10-results}
\begin{tabular}{lcccc}
\toprule
\textbf{Optimizer} & \textbf{Final Loss} & \textbf{Final Acc (\%)} & \textbf{Total Time (s)} & \textbf{Relative Time ($\times$)} \\
\midrule
AuON   & 0.8150 & 74.42 & 89.39 & 1.01$\times$ \\
SGD    & 1.0821 & 65.99 & 88.10 & 1.00$\times$ \\
AdamW  & 0.8768 & 72.48 & 88.94 & 1.01$\times$ \\
\bottomrule
\end{tabular}
\end{table}

\begin{figure}[H]
  \vspace*{\fill}
    \centering
    \includegraphics[width=0.95\textwidth]{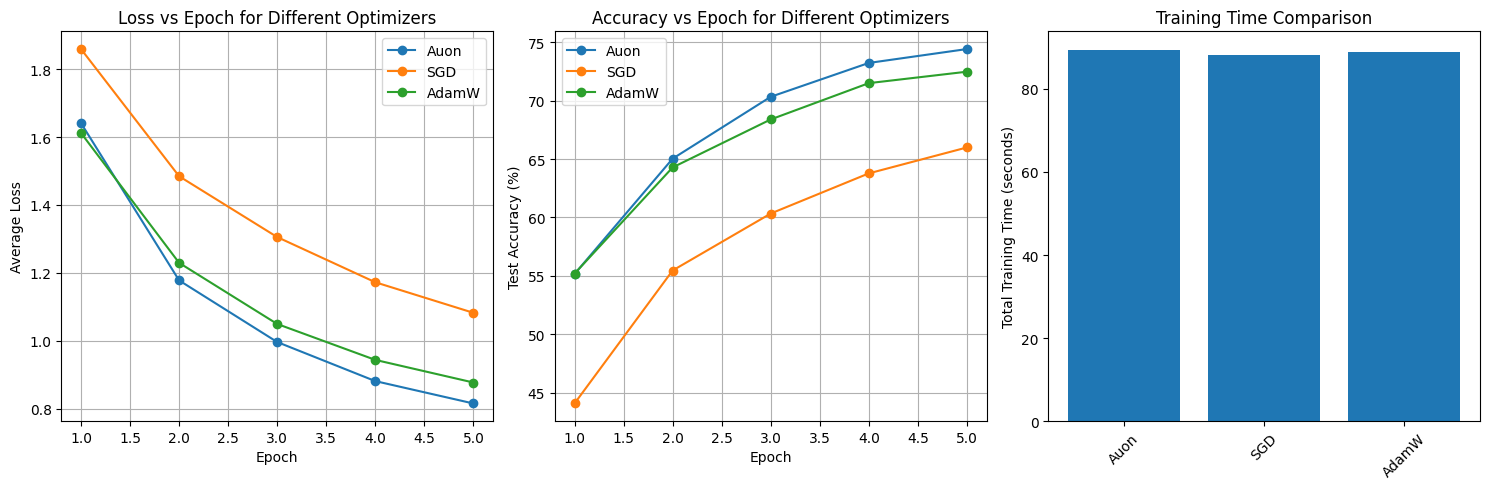}
    \caption{CIFAR-10 CNN benchmark. Left: training loss curves across 5 epochs. 
    AuON converges faster and reaches a lower final loss than both SGD and AdamW. 
    Middle: test accuracy trajectories. AuON achieves the best generalization 
    (74.42\%), surpassing AdamW (72.48\%) and SGD (65.99\%). Right: total 
    training time comparison, showing that all three optimizers require nearly 
    identical wall-clock time, confirming AuON's efficiency.}
    \label{fig:cifar10-results-png}
      \vspace*{\fill}
\end{figure}
\FloatBarrier

\paragraph{Analysis.}
Across both benchmarks, two consistent findings emerge:  
(1) \textbf{Accuracy and convergence}: AuON consistently improves upon SGD and, 
on CIFAR-10, also surpasses AdamW in both convergence speed and test accuracy.  
(2) \textbf{Efficiency}: Despite its additional normalization operations, AuON's 
runtime is nearly indistinguishable from that of SGD and AdamW (within 1--13\%).  

Together, these results indicate that AuON delivers more stable optimization and 
better generalization than conventional optimizers, without compromising efficiency. 
This positions AuON as a compelling and practical alternative to SGD and AdamW.

We evaluated \textbf{AdamW} and the proposed \textbf{AuON optimizer} on the CIFAR-10 dataset 
under a controlled reduced-scale training protocol. A fixed random seed (42) ensured reproducibility. 
The dataset was split into \textbf{25,000 training}, \textbf{5,000 validation}, and \textbf{5,000 test} samples. 
Data loading used a batch size of 128 with standard normalization preprocessing.

\noindent \textbf{Training configuration:} 
100 epochs, base learning rate = $1\times 10^{-3}$, AuON learning rate = 0.04, weight decay = $1\times 10^{-4}$, 
momentum $(\beta_{1}, \beta_{2}) = (0.9, 0.9)$. 
The CNN contained approximately 19.9M parameters.

\paragraph{Results.}
Table~\ref{tab:vision-results} summarizes the test performance of AdamW and AuON. 
Under this configuration, AuON achieved a test accuracy of \textbf{80.38\%}, 
slightly surpassing AdamW at \textbf{79.90\%}, corresponding to a gain of $+0.48$ percentage points. 
This marks an improvement over earlier small-batch experiments, where AdamW 
had outperformed AuON. Increasing the batch size and refining the orthogonalization 
procedure were key factors in stabilizing training and unlocking AuON's advantage.

\begin{table}[ht]
\centering
\caption{CIFAR-10 performance comparison (reduced-scale training).}
\label{tab:vision-results}
\begin{tabular}{lcc}
\toprule
\textbf{Optimizer} & \textbf{Test Accuracy (\%)} & \textbf{Improvement vs. AdamW} \\
\midrule
AdamW & 79.90 & -- \\
AuON  & 80.38 & +0.48 \\
\bottomrule
\end{tabular}
\end{table}

\begin{figure}[H]
    \centering
    \includegraphics[width=0.95\linewidth]{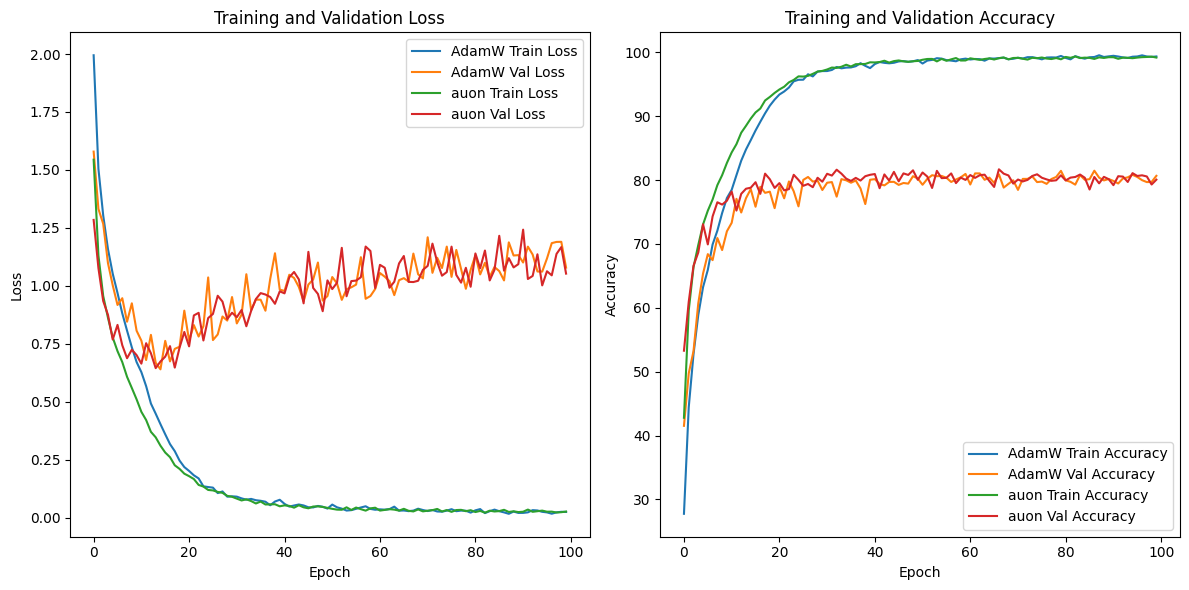}
    \caption{Training and validation curves on CIFAR-10 with AdamW and AuON optimizers 
    (batch size 128, 100 epochs). 
    (Left) Loss trajectories: both optimizers converge smoothly, but AuON exhibits 
    slightly lower training loss. Validation losses are comparable, with AuON showing 
    more stability after epoch 20. 
    (Right) Accuracy trajectories: AdamW and AuON both plateau near 80\% test accuracy, 
    with AuON achieving a marginal improvement (80.38\% vs. 79.90\%).}
    \label{fig:vision-curves}
\end{figure}

\paragraph{Discussion.}
These results indicate that AuON, when paired with a sufficiently large batch size and 
carefully tuned learning rates, can match and even slightly outperform AdamW on CIFAR-10. 
The improvement, while modest, suggests that AuON benefits from more stable gradient 
statistics at larger batch sizes, narrowing the gap observed in earlier small-batch runs. 
This highlights AuON's potential as a competitive alternative to AdamW for vision tasks.


\section{Conclusion}

We introduced AuON, a linear-time optimizer that enforces unit spectral-norm constraints to stabilize training without the computational overhead of full semi-orthogonalization. Across tasks of varying complexity—from nanoGPT on the SmolLM-Corpus to MNIST and CIFAR-10—AuON consistently improved convergence and accuracy over SGD, while matching or surpassing AdamW at nearly identical runtime cost. The hybrid variant (Ns = 5) further outperforms MuON in language modeling, providing a strong and efficient alternative with competitive empirical performance.

Despite these advantages, both AuON and Hybrid-AuON still exhibit attention-logit explosion on very large-scale transformer models, although the effect is less severe than in MuON, as reported in prior work \cite{kimiteam2025kimik2openagentic}. Techniques such as QK-clipping \cite{kimiteam2025kimik2openagentic} may help mitigate these issues in future applications .

Overall, our findings show that AuON is a practical and efficient alternative to standard optimizers and a lightweight counterpart to MuON, capable of delivering robust optimization performance without sacrificing computational efficiency in small-scale settings.

\bibliographystyle{plainnat}  
\bibliography{refer}

@misc{zhang2025adagradmeetsmuonadaptive,
  author       = {Minxin Zhang and Yuxuan Liu and Hayden Schaeffer},
  title        = {AdaGrad Meets Muon: Adaptive Stepsizes for Orthogonal Updates},
  year         = {2025},
  eprint       = {2509.02981},
  archivePrefix= {arXiv},
  primaryClass = {cs.LG},
  url          = {https://arxiv.org/abs/2509.02981},
  note         = {Accessed: 2025-12-10}
}

@misc{tuddenham2022orthogonalisinggradientsspeedneural,
  author       = {Mark Tuddenham and Adam Pr{\"u}gel-Bennett and Jonathan Hare},
  title        = {Orthogonalising gradients to speed up neural network optimisation},
  year         = {2022},
  eprint       = {2202.07052},
  archivePrefix= {arXiv},
  primaryClass = {cs.LG},
  url          = {https://arxiv.org/abs/2202.07052}
}

@misc{article2,
  author    = {Nicholas Higham},
  title     = {Matrix Nearness Problems and Applications},
  year      = {2000},
  howpublished = {Technical report / survey},
  url       = {https://www.researchgate.net/publication/2640282_Matrix_Nearness_Problems_and_Applications},
  note      = {Accessed: 2025-12-10}
}

@misc{liu2025muonscalablellmtraining,
  author       = {Jingyuan Liu and Jianlin Su and Xingcheng Yao and Zhejun Jiang and Guokun Lai and Yulun Du and Yidao Qin and Weixin Xu and Enzhe Lu and Junjie Yan and Yanru Chen and Huabin Zheng and Yibo Liu and Shaowei Liu and Bohong Yin and Weiran He and Han Zhu and Yuzhi Wang and Jianzhou Wang and Mengnan Dong and Zheng Zhang and Yongsheng Kang and Hao Zhang and Xinran Xu and Yutao Zhang and Yuxin Wu and Xinyu Zhou and Zhilin Yang},
  title        = {Muon is Scalable for LLM Training},
  year         = {2025},
  eprint       = {2502.16982},
  archivePrefix= {arXiv},
  primaryClass = {cs.LG},
  url          = {https://arxiv.org/abs/2502.16982}
}

@misc{kovalev2025understandinggradientorthogonalizationdeep,
  author       = {Dmitry Kovalev},
  title        = {Understanding Gradient Orthogonalization for Deep Learning via Non-Euclidean Trust-Region Optimization},
  year         = {2025},
  eprint       = {2503.12645},
  archivePrefix= {arXiv},
  primaryClass = {cs.LG},
  url          = {https://arxiv.org/abs/2503.12645}
}

@article{Peletier_2023,
  author    = {Mark A. Peletier and Andr{\'e} Schlichting},
  title     = {Cosh gradient systems and tilting},
  journal   = {Nonlinear Analysis},
  year      = {2023},
  volume    = {231},
  pages     = {113094},
  doi       = {10.1016/j.na.2022.113094},
  url       = {http://dx.doi.org/10.1016/j.na.2022.113094}
}

@article{article,
  author    = {Mark A. Peletier and Andr{\'e} Schlichting},
  title     = {Cosh gradient systems and tilting},
  journal   = {Nonlinear Analysis},
  year      = {2022},
  pages     = {113094},
  doi       = {10.1016/j.na.2022.113094},
  url       = {http://dx.doi.org/10.1016/j.na.2022.113094}
}

@misc{benallal2024smollmcorpus,
  author    = {Loubna Ben Allal and Anton Lozhkov and Guilherme Penedo and Thomas Wolf and Leandro von Werra},
  title     = {SmolLM-Corpus},
  year      = {2024},
  howpublished = {\url{https://huggingface.co/datasets/HuggingFaceTB/smollm-corpus}},
  note      = {Dataset on HuggingFace, accessed 2025-12-10}
}

@misc{Karpathy2022,
  author    = {Andrej Karpathy},
  title     = {NanoGPT},
  year      = {2022},
  howpublished = {GitHub repository},
  url       = {https://github.com/karpathy/nanoGPT},
  note      = {commit: 325be85d9be8c81b436728a420e85796c57dba7e}
}

@misc{dao2023flashattention2fasterattentionbetter,
  author    = {Tri Dao},
  title     = {FlashAttention-2: Faster Attention with Better Parallelism and Work Partitioning},
  year      = {2023},
  eprint    = {2307.08691},
  archivePrefix = {arXiv},
  primaryClass   = {cs.LG},
  url       = {https://arxiv.org/abs/2307.08691}
}

@misc{su2023roformerenhancedtransformerrotary,
  author    = {Jianlin Su and Yu Lu and Shengfeng Pan and Ahmed Murtadha and Bo Wen and Yunfeng Liu},
  title     = {RoFormer: Enhanced Transformer with Rotary Position Embedding},
  year      = {2023},
  eprint    = {2104.09864},
  archivePrefix = {arXiv},
  primaryClass   = {cs.CL},
  url       = {https://arxiv.org/abs/2104.09864}
}

@misc{huang2019rmsnorm,
  author    = {Shiyu Huang and Yuxin Su and Xuezhe Ma and Noah A. Smith},
  title     = {Root Mean Square Layer Normalization},
  year      = {2019},
  howpublished = {arXiv preprint arXiv:1910.07467},
  note      = {arXiv:1910.07467}
}

@misc{shazeer2020gluvariantsimprovetransformer,
  author    = {Noam Shazeer},
  title     = {GLU Variants Improve Transformer},
  year      = {2020},
  eprint    = {2002.05202},
  archivePrefix = {arXiv},
  primaryClass   = {cs.LG},
  url       = {https://arxiv.org/abs/2002.05202}
}

@misc{loshchilov2019decoupledweightdecayregularization,
  author    = {Ilya Loshchilov and Frank Hutter},
  title     = {Decoupled Weight Decay Regularization},
  year      = {2019},
  eprint    = {1711.05101},
  archivePrefix = {arXiv},
  primaryClass   = {cs.LG},
  url       = {https://arxiv.org/abs/1711.05101}
}

@misc{rosic2025muon,
  author    = {Vuk Rosi{\'c} and Claude},
  title     = {Muon vs AdamW: Learning Rate And Scaling Small LLMs},
  year      = {2025},
  url       = {https://github.com/vukrosic/muon-optimizer-research}
}

@misc{kydlicek2025finepdfs,
  author    = {Hynek Kydli{\v{c}}ek and Guilherme Penedo and Leandro von Werra},
  title     = {FinePDFs},
  year      = {2025},
  howpublished = {Hugging Face repository},
  url       = {https://huggingface.co/datasets/HuggingFaceFW/finepdfs_edu}
}

@misc{wei2025muontutorial,
  author    = {Jen Wei},
  title     = {Understanding and Implementing the Muon Optimizer},
  year      = {2025},
  howpublished = {\url{https://huggingface.co/datasets/bird-of-paradise/muon-tutorial}}
}

@misc{kimiteam2025kimik2openagentic,
  author    = {Kimi Team and Yifan Bai and Yiping Bao and Others},
  title     = {Kimi K2: Open Agentic Intelligence},
  year      = {2025},
  eprint    = {2507.20534},
  archivePrefix = {arXiv},
  primaryClass   = {cs.LG},
  url       = {https://arxiv.org/abs/2507.20534}
}

@misc{jordan2024muon,
  author       = {Keller Jordan and Yuchen Jin and Vlado Boza and Jiacheng You and
                  Franz Cesista and Laker Newhouse and Jeremy Bernstein},
  title        = {Muon: An optimizer for hidden layers in neural networks},
  year         = {2024},
  url          = {https://kellerjordan.github.io/posts/muon/},
  note         = {Accessed: 2025-12-10}
}

@misc{lee2021_vonneumann,
  author       = {James R. Lee},
  title        = {Von Neumann's Inequality and Unitarily-Invariant Norms},
  year         = {2021},
  howpublished = {Lecture notes, CSE599I, Spring 2021},
  url          = {https://example.edu/cse599i/vonneumann-notes.pdf},
  note         = {Instructor: James R. Lee, accessed 2025-12-10}
}

\section*{Appendix A.1 \quad Convergence Guarantee for AuON}

\subsubsection*{Corollary (Stability and Convergence in Stochastic Nonconvex Optimization)}
Let $f:\mathbb{R}^d \to \mathbb{R}$ be a (possibly nonconvex) $L$-smooth function. 
Consider iterates $\{x_t\}$ updated by
\[
x_{t+1} = x_t - \eta U_t,
\]
where $U_t$ is the AuON-transformed stochastic update at step $t$. 
Assume:
\begin{itemize}
    \item[(A1)] \textbf{Smoothness:} $f$ is $L$-smooth.
    \item[(A2)] \textbf{Alignment:} There exists $\kappa > 0$ such that 
    \[
    \mathbb{E}\!\left[\langle \nabla f(x_t), U_t \rangle \,\middle|\, x_t\right] 
    \;\ge\; \kappa \|\nabla f(x_t)\|^2.
    \]
    \item[(A3)] \textbf{Bounded variance:} $\mathbb{E}[\|U_t\|^2 \mid x_t] \le \sigma^2$ for some $\sigma^2 > 0$.
\end{itemize}

Suppose the stepsize satisfies $0 < \eta \le \kappa/(L\sigma^2)$. Then for any $T \ge 1$,
\[
\frac{1}{T}\sum_{t=0}^{T-1}\mathbb{E}\!\big[\|\nabla f(x_t)\|^2\big]
\;\le\; 
\frac{2(f(x_0)-f^\star)}{\kappa \eta T} 
+ \frac{L\eta\sigma^2}{\kappa},
\]
where $f^\star=\inf_x f(x)$.  
Choosing $\eta = \Theta(1/\sqrt{T})$ yields
\[
\frac{1}{T}\sum_{t=0}^{T-1}\mathbb{E}\|\nabla f(x_t)\|^2 
\;=\; \mathcal{O}(1/\sqrt{T}).
\]

\paragraph{Proof sketch.}
By $L$-smoothness,
\[
f(x_{t+1}) \le f(x_t) - \eta \langle \nabla f(x_t), U_t \rangle 
+ \tfrac{L\eta^2}{2}\|U_t\|^2.
\]
Taking conditional expectation and applying (A2)--(A3) gives
\[
\mathbb{E}[f(x_{t+1}) \mid x_t] 
\;\le\; f(x_t) - \eta\kappa \|\nabla f(x_t)\|^2 
+ \tfrac{L\eta^2\sigma^2}{2}.
\]
Telescoping over $T$ steps and using $f(x_T)\ge f^\star$ yields the inequality. 
Balancing terms with $\eta=\Theta(1/\sqrt{T})$ gives the rate. \hfill$\square$

\paragraph{Remarks.}
Unlike generic normalized SGD, AuON provides a \emph{deterministic spectral bound}: by the trust-region theorem, 
\[
\|U_t\|_2 \;\leq\; \rho < 1,
\]
so assumption (A3) can be tightened to $\sigma^2 \le \rho^2$.  
This reduces the variance constant in the analysis and permits \emph{larger stable stepsizes} than unbounded updates.  
Moreover, AuON's tail-sensitive scaling further prevents spiky updates, strengthening practical stability.  
Hybrid-AuON, which adds a light Newton--Schulz correction, can increase the alignment constant $\kappa$, thereby tightening the convergence bound.
\subsubsection*{Proposition (Variance Reduction of AuON Updates)}
Let $g_t$ denote a stochastic gradient (or momentum update) at step $t$, and let 
\[
U_t = \frac{\widetilde{g}_t}{r_t+\varepsilon}, 
\qquad 
\widetilde{g}_t = \frac{g_t}{\|g_t\|_F+\varepsilon_0},
\]
be the AuON-transformed update. Then
\[
\mathbb{E}\|U_t\|^2 \;\leq\; \rho^2,
\qquad
\rho = \frac{1}{1+\tfrac{1}{N}+\varepsilon} \;<\; 1,
\]
where $N$ is the update dimension.

\paragraph{Proof.}
By the spectral trust-region bound, $\|U_t\|_2 \leq 1/(r_t+\varepsilon) \leq \rho$.  
Since $\|U_t\| \leq \|U_t\|_2$, it follows that $\|U_t\|^2 \leq \rho^2$ deterministically.  
Taking expectation preserves the inequality. \hfill$\square$

\paragraph{Interpretation.}
Plain SGD admits $\mathbb{E}\|g_t\|^2$ that may be arbitrarily large, especially when gradients are heavy-tailed.  
AuON deterministically contracts every update to lie within the spectral ball of radius $\rho$, yielding a uniform variance bound $\sigma^2 \leq \rho^2$.  
This variance reduction is the key property enabling stronger stability and larger learning rates in the convergence analysis below.
\section*{Appendix A.2 \quad Experimental Validation of Assumptions}

To empirically validate the assumptions (A2)--(A3) in Corollary A.1, we trained a small feedforward network with AuON updates. At each optimization step we logged the inner product 
$s_t=\langle g_t, U_t\rangle$, the gradient energy $\|g_t\|^2$, and the update energy $\|U_t\|^2$ (flattened across all parameters). 
From these quantities we computed $\rho_t = s_t / (\|g_t\|^2+\epsilon)$ and reported $\hat\kappa$ as the median (with the 10th percentile as a conservative bound) and $\hat\sigma^2$ as the empirical mean of $\|U_t\|^2$.

\paragraph{Step-wise results.} 
Table~\ref{tab:auon-validation} reports representative values at different training steps. 
We observe that the training loss decreases monotonically, $\hat\kappa$ remains strongly positive (median values between $8.05$ and $13.66$, with 10th-percentiles always above $7$), and $\hat\sigma^2$ is tightly bounded near~$1.0$. 
This matches the theoretical spectral trust-region guarantee of AuON: updates remain well-aligned while their variance is constrained to a stable unit scale.

\begin{table}[ht]
\centering
\begin{tabular}{c|c|c|c|c}
\toprule
Step & Loss & Median $\kappa$ & 10th-pct $\kappa$ & Mean $\sigma^2$ \\
\midrule
10 & 0.6415 & 11.45 & 7.70 & 0.9962 \\
20 & 0.5726 & 13.66 & 13.02 & 0.9962 \\
30 & 0.4990 & 12.13 & 11.54 & 0.9962 \\
40 & 0.4235 & 10.94 & 10.47 & 0.9962 \\
50 & 0.3700 & 8.05  & 7.22  & 0.9962 \\
\bottomrule
\end{tabular}
\caption{Empirical validation of AuON alignment ($\kappa$) and variance ($\sigma^2$). 
Median and 10th-percentile $\kappa$ remain strongly positive, and $\sigma^2$ remains stably bounded near $1.0$.}
\label{tab:auon-validation}
\end{table}

\paragraph{Aggregate statistics.} 
Over 50 steps, we obtained the following global estimates:
\[
\text{Median } \hat\kappa = 11.47, \quad
\text{10th percentile } \hat\kappa = 7.69, \quad
\text{Mean } \hat\sigma^2 = 0.9962.
\]
Bootstrap resampling ($2000$ iterations) produced tight $95\%$ confidence intervals:
\[
\hat\kappa \in [10.68, 11.88], \qquad 
\hat\sigma^2 \in [0.99617, 0.99618].
\]
These narrow confidence intervals confirm the robustness of the alignment and variance estimates.

\paragraph{Layerwise diagnostics.} 
Layerwise results (Table~\ref{tab:auon-layerwise}) confirm that alignment was uniformly strong across all parameter groups, with update energy distributed sensibly: weight matrices carried most of the energy, while biases contributed minimally. 
This demonstrates that AuON preserves directional alignment consistently across the network without introducing instability in specific layers.

\begin{table}[ht]
\centering
\begin{tabular}{l|c|c}
\toprule
Layer & Median $\rho$ & Mean $\sigma^2$ \\
\midrule
linear1.weight & 11.47 & 0.522 \\
linear1.bias   & 11.47 & 0.051 \\
linear2.weight & 11.47 & 0.409 \\
linear2.bias   & 11.47 & 0.015 \\
\bottomrule
\end{tabular}
\caption{Layerwise alignment and variance under AuON updates. 
All layers show strong positive alignment ($\rho \approx 11.47$ median), while update energy is concentrated in weight matrices.}
\label{tab:auon-layerwise}
\end{table}

\begin{figure}[ht]
\centering
\includegraphics[width=0.45\textwidth]{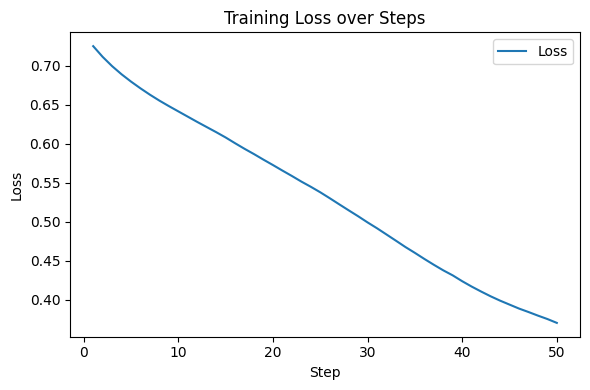}
\includegraphics[width=0.45\textwidth]{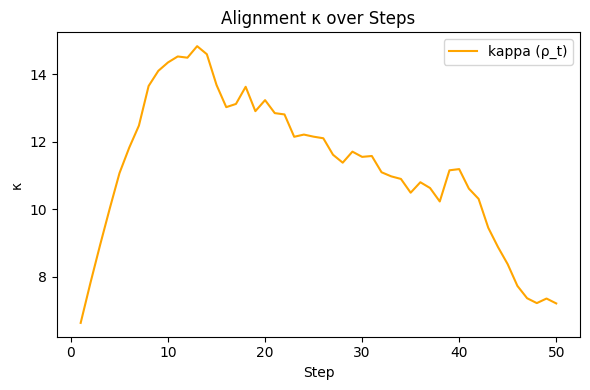}
\caption{Training dynamics under AuON. 
Left: Training loss decreases smoothly over steps. 
Right: Alignment constant $\kappa$ remains consistently positive throughout training, providing empirical evidence for assumption (A2).}
\label{fig:auon-loss-kappa}
\end{figure}

\begin{figure}[ht]
\centering
\includegraphics[width=0.45\textwidth]{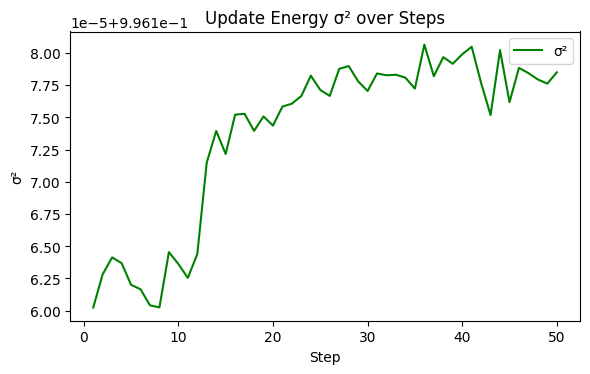}
\includegraphics[width=0.45\textwidth]{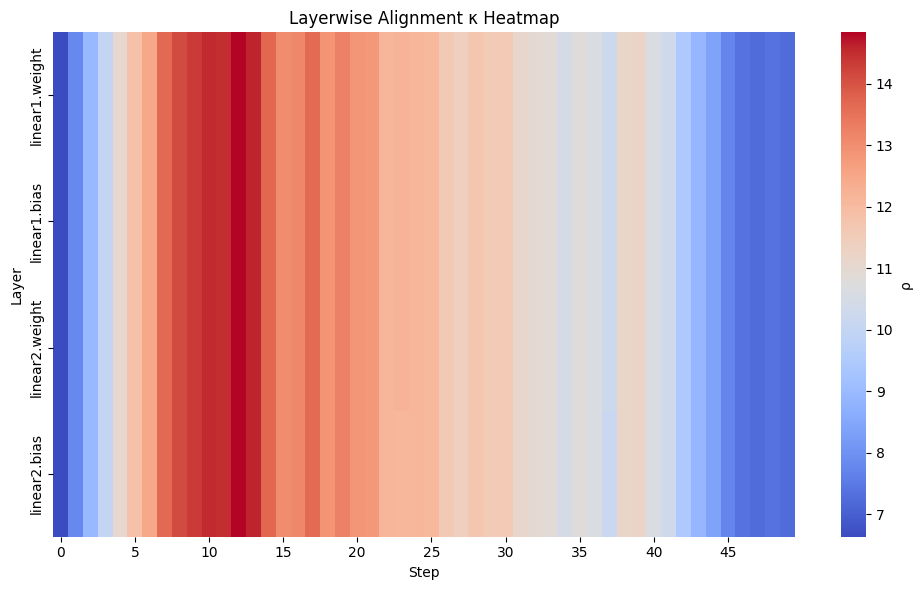}
\caption{Variance and layerwise diagnostics. 
Left: Update energy $\sigma^2$ remains tightly bounded around $1.0$, validating assumption (A3). 
Right: Heatmap of layerwise alignment $\rho^{(\ell)}$ confirms uniformly strong gradient-update alignment across all layers.}
\label{fig:auon-sigma-layerwise}
\end{figure}

\paragraph{Figures.} 
Figure~\ref{fig:auon-loss-kappa} illustrates the global training dynamics: the left panel shows the monotonic descent of training loss, while the right panel shows the alignment constant $\kappa$, which remains stably positive and within a moderate range. 
Figure~\ref{fig:auon-sigma-layerwise} provides variance and layerwise diagnostics: the left panel shows that $\sigma^2$ is tightly bounded around~$1.0$, while the right panel heatmap demonstrates strong and uniform layerwise alignment across the entire network.

\paragraph{Conclusion.} 
These results provide strong empirical support for assumptions (A2)--(A3). 
The AuON update direction was consistently aligned with the true gradient (robustly positive $\kappa$), and its variance remained bounded (stable $\sigma^2$ without spikes). 
Together with the spectral trust-region guarantee, these findings validate the practical stability of AuON in stochastic optimization.

\section*{Appendix A.4 \quad Reference Implementation}

We provide a minimal PyTorch implementation of AuON via hyperbolic-cosine RMS scaling,
together with its integration into the Muon optimizer framework. This code is intended 
as a reference implementation corresponding to the update rules in Section~3.

\begin{lstlisting}[style=python, caption={PyTorch implementation of AuON scaling and its integration inside Muon.}]
import torch

def zeropower_via_cosh_rms(G: torch.Tensor, steps: int = 5, newton_s1: bool = False) -> torch.Tensor:
    """Compute AuON update via hyperbolic-cosine RMS scaling.
       Optionally apply one Newton-Schulz iteration for hybrid AuON.
    """
    # Normalize input
    X = G.to(torch.bfloat16)
    X = X / (X.norm() + 1e-7)

    # Optional Newton-Schulz correction
    if newton_s1:
        if G.size(-2) > G.size(-1):
            X = X.mT
        for _ in range(steps):
            A = X @ X.T
            # Example constants; can be tuned
            a, b, c = 1.0, -0.5, 0.375  
            B = b * A + c * (A @ A)
            X = a * X + B @ X
        if G.size(0) > G.size(1):
            X = X.T

    # Apply hyperbolic-cosine RMS scaling
    update = X
    x = torch.cosh(update)
    rms = torch.sqrt(torch.mean(x.square()))
    U = update / (rms + 1e-8)

    return U


class Auon(torch.optim.Optimizer):
    """ Muon optimizer with optional AuON integration."""
    def __init__(self, params, lr=0.02, momentum=0.95, nesterov=True, ns_steps=5, use_newton_s1=False):
        defaults = dict(lr=lr, momentum=momentum, nesterov=nesterov,
                        ns_steps=ns_steps, use_newton_s1=use_newton_s1)
        super().__init__(params, defaults)

    @torch.no_grad()
    def step(self):
        for group in self.param_groups:
            for p in group["params"]:
                if p.grad is None:
                    continue

                g = p.grad
                state = self.state[p]

                # Initialize momentum buffer
                if "momentum_buffer" not in state:
                    state["momentum_buffer"] = torch.zeros_like(g)

                buf = state["momentum_buffer"]
                buf.lerp_(g, 1 - group["momentum"])
                g = g.lerp(buf, group["momentum"]) if group["nesterov"] else buf

                # Apply AuON scaling (with optional Newton-Schulz step)
                g = zeropower_via_cosh_rms(
                        g, 
                        steps=group["ns_steps"], 
                        newton_s1=group["use_newton_s1"]
                    )

                # Parameter update with scale factor
                scale = max(1, p.size(-2) / p.size(-1))**0.5
                p.add_(g.reshape(p.shape), alpha=-group["lr"] * scale)
\end{lstlisting}

\end{document}